\pdfoutput=1
\documentclass[11pt]{article}

\usepackage{ACL2023}
\usepackage{microtype}
\usepackage{times}
\usepackage{latexsym}
\usepackage[utf8]{inputenc}
\usepackage{newunicodechar}
\newunicodechar{−}{-}
\usepackage[T1]{fontenc}
\usepackage{wrapfig}
\usepackage{microtype}

\usepackage{inconsolata}

\usepackage{latexsym}
\usepackage{makecell}
\usepackage{graphicx}
\graphicspath{{figures/}}
\usepackage{amsmath}
\usepackage{pifont}% 
\usepackage{hyperref}
\usepackage{makecell}
\title{Computational Sentence-level Metrics for Predicting Comprehension of Entire Sentence by Humans}
\author{Kun Sun \\
 University of Tübingen \\
  \texttt{kun.sun@uni-tuebingen.de}  \\\And
  Rong Wang \\ University of Stuttgart, Germany\\
  	\texttt{rongw.de@gmail.com}\\
}
\begin{document}
	
\maketitle
\begin{abstract}
	The majority of research in computational psycholinguistics has concentrated on the processing of words. This study introduces innovative methods for computing sentence-level metrics using multilingual large language models. The metrics developed—\textit{sentence surprisal} and \textit{sentence relevance}— are tested and compared to validate whether they can predict how humans comprehend sentences as a whole across languages.
%This study proposes a novel approach to predicting how human comprehend sentences as a whole, using \texttt{sentence surprisal} and \texttt{sentence relevance} measures computed with multilingual transformer-based language models across multiple languages. The present research introduces attention-aware methods inspired by the attention mechanism utilized in Transformers. %These methods compute sentence-level surprisal by taking into account the surprisal values of neighboring sentences. 
%In order to appraise the baseline metric, this study carries out a comparison and examination of three distinct sentence-level metrics. 
%The evaluation results demonstrate that the attenion-aware measures, which incorporates contextual information fully and comprehensively, outperform the baseline metric. 
These metrics offer significant interpretability and achieve high accuracy in predicting human sentence reading speed. Our results indicate that these computational sentence-level metrics are exceptionally effective at predicting and elucidating the processing difficulties encountered by readers in comprehending sentences as a whole across a variety of languages.
%These metrics are highly interpretable, and acheive high accuracy in predicting human sentence reading speed. Our findings reveal that these computational metrics are highly capable of predicting and explaining the processing difficulty of readers comprehending sentences as a whole for multiple languages. 
Their impressive performance and generalization capabilities provide a promising avenue for future research in integrating LLMs and cognitive science. 
\end{abstract}

\pagestyle{plain} % Just a simple page number

\section{Introduction}

 In the study of how humans comprehend and process language, computational models play a crucial role in understanding the connections between linguistic features and behavior/neural signals. These models can be used to make linguistic predictions, model language features, or specify the processing steps that can be quantitatively compared to behavioral and neural signals.% Prediction is a crucial factor in both language comprehension and processing (\citealp{clark2013whatever};  \citealp{kuperberg2016we}). 
 With the abundance of experimental data available, researchers have developed several computational models/metrics to simulate how humans comprehend a given linguistic unit within a given context. 
 
 One such model measures the information communicated by any particular linguistic component, whether it is a phoneme, a word, or an entire utterance, when considered within its left context. This method is commonly known as \textbf{surprisal} \cite{crocker2016information}. %One such model estimates word prediction by calculating the transitional probability from the current word to the next word, 
\textbf{Word surprisal} estimates the information among words in context, and this metric has proven to be influential in predicting human word processing (\citealp{hale2001probabilistic}; \citealp{levy2008expectation}). In contrast to \textbf{expectation-based} models, \textbf{memory-based} models rely on a memory mechanism to store and retrieve information from previous input (e.g., ACT-R (\citealp{r1975computer}, short-term memory \citep{baddeley2010working}). %These models retrieve information from memory based on the current context and can use this information to understand the meaning of the text in terms of a previously established semantic relatedness. 
 %For example, ACT-R (\citealp{r1975computer}) and short-term memory (\citealp{baddeley2010working}) are two examples of memory-based models. %Deep learning models, such as RNN and Transformer, have been inspired by memory-based models to some extent. 
 One metric used to estimate language processing involves assessing the semantic relation with other words, such as through \textbf{semantic similarity} (\citealp{mitchell2010composition}; \citealp{hollis2016principals}). 

%In the following section, we provide a detailed account of two main computational theories used to predict language comprehension and processing.
 %\texttt{Surprisal} can be taken as the expectation-based metric, which suggests that processing difficulty is largely determined by how predictable the upcoming linguistic material is in its context. In cognitive modeling, predictability is often operationalized by information-theoretic surprisal (\citealp{shannon1948mathematical}), which has been shown to be a strong predictor of behavioral and neural processing measures of processing difficulty.
 
  \texttt{Word surprisal}, an expectation-based metric, provides empirical supports for the position that words are more difficult to process when they are harder to anticipate from preceding context (\citealp{demberg2008data}; \citealp{smith2013effect}; \citealp{hale2016information}; \citealp{shain2020fmri}). However, recent work has shown that surprisal computed by neural language models tends to underpredict human reading times of both targeted constructions and naturalistic text (\citealp{van2021single}; \citealp{arehalli2022syntactic}; \citealp{oh2023does}). % and they also demonstrated that neural LMs yield surprisal estimates that underpredict naturalistic reading times of English and Japanese text compared to those from neural LMs that have a recency bias implemented as limited access to the previous context.
   \textit{Does this imply that surprisal has lost its efficacy in predicting reading time?} Perhaps it is time to reevaluate this conventional metric.
 % Grounded in the principles of information theory, \textit{surprisal} unifies the notions of incremental parsing and expectations into a single account. Due to the mechanisms of prediction and integration, predictable words will be easier to process and cognitive effort required to process a word is proportional to its surprisal: difficulty(w$^$t) ≈ −log $P(wt|w1 … wt−1, CONTEXT)$. However, it remains an open question, what aspects of the context – lexical, syntactic, semantic, conceptual – are used for prediction or to facilitate integration? If CONTEXT denotes extra-sentential context, it can go beyond the current sentence being analyzed and include information from previous sentences or from the broader discourse context in which the sentence occurs. But there is few computational models take account of the aspects and scopes of context.  
%On the other hand, the memory-based theory suggests that processing difficulty stems from limitations in the ability to store representations of the preceding context and to retrieve and integrate them with new input. For example, the dependency-locality measure is a memory-based model that has been useful in explaining how some special syntactic structures are processed (\citealp{gibson1998linguistic}; \citealp{vasishth2006argument}). However, this measure has not been as successful in predicting how words are processed in context by humans as surprisal. 
On the other hand, the memory-based theory posits processing difficulty arises from storing, retrieving, and integrating previous context with new input. For example, while the dependency-locality model explains certain syntactic structures processing (\citealp{gibson1998linguistic}; \citealp{vasishth2006argument}), it has not been as effective as surprisal in predicting humans' contextual word processing. Another memory-based metric, semantic similarity, gauges the similarity between the meanings of two words or phrases and is effective in predicting how words are processed. Contextual semantic similarity, which was developed from semantic similarity, concerns the semantic relatedness between a target word and its contextual words. Substantial empirical evidence supports the effects of word-level semantic similarity in language comprehension (\citealp{roland2012semantic}; \citealp{broderick2018electrophysiological}; \citealp{sun2023interpretable}; \citealp{sun2023optimizing}).

Although the related studies have claimed that some given computational metrics are predictive of sentence processing by humans, the fact is that these metrics can only predict how words are processed in a given context (such as  \texttt{word surprisal}) (\citealp{hale2016information}; \citealp{de2023scaling}; \citealp{gotlieb2023testing}). Focusing solely on word processing may not truly mirror the real-world human experience of comprehending sentences. In other words, we should also explore how humans comprehend entire sentences, rather than merely the individual words within them. It is necessary to broaden our understanding of sentence comprehension, placing human sentence processing within a larger context: discourse. Computational sentence-level metrics (such as sentence probability, sentence semantic relevance) are potentially useful in understanding entire sentence comprehension and processing by humans. %Sentence-level computational metrics can predict how humans process a sentence as a whole instead of just focusing on words within a sentence. 
For instance, the metric of contextual sentence-level relevance computed by Transformer-based language models could predict how Chinese natives comprehend sentences as a whole \citep{sun2022semantic}. A similar method can be applied in investigating sentence comprehension in other languages. %With the wide availability of experimental data and recent developments in NLP, we are highly likely to precisely compute next sentence probabilities and sentence (textual) similarity.
In short, we can enhance our analysis from word-level metrics to sentence-level metrics, exploring their predictability for the processing of a sentence within a discourse-level context.    %For instance, we can use ``multilingual-BERT'' (\texttt{m-BERT}) \citep{devlin2018bert} to compute the sentence probabilities and sentence similarities for a variety of languages, and thus to then predict how humans comprehend sentence across a range of different languages. 

During naturalistic discourse reading, a number of factors may have an effect on sentence processing, including expectations concerning the next sentence based on the preceding context, and memory integrated with new input based on semantic relatedness with the context. Surprisal reflects the amount of information conveyed by a linguistic unit and its predictability, while semantic relevance represents the relatedness of a linguistic unit to the other units. Both factors could independently influence processing difficulty, but they may also interact, creating complex processing situations. The study aims to compute these metrics at the senentece level and test hypotheses about their nature and understand their potentials in human sentence comprehension.
With the advent of deep learning techniques and the availability of massive datasets, there is now an opportunity to develop and compute such sentence-level metrics. %The current study will investigate how the computational metrics we propose can predict sentence processing by leveraging the power of multilingual, pre-trained language models such as \texttt{m-BERT}. And
By using multi-lingual large language models (LLMs), we can estimate the probabilities of the next sentence (\textbf{sentence surprisal}) or compute semantic relevance among sentences (\textbf{sentence relevance}). These sentence-level metrics we proposed are expected to predict how humans comprehend and process sentence as a whole. Because of sentence length and human memory-stored limit, we consider a limited number of surrounding sentences as the context in calculating contextual information and further incorporate the contextual information in our metrics. The method is ``attention-aware'' approach because it works as effectively as the attention mechanism in Transformer (\citealp{vaswani2017attention}) in processing contextual information. %Sentence relevance computed by the attention-aware approach is expected to better predict how human beings processes sentences as a whole. %Furthermore, the combination of sentence surprisal and sentence relevance can result in an additional measure, which we refer to as the ``enhanced measure'' and use to make predictions in human sentence comprehension.

We plan to apply these computational metrics to test their predictability in a number of languages rather than in English. %In other words, we will carry out a multi-lingual investigation to test the universal significance of these measures. However, such cross-lingual investigations would need to draw on  multi-lingual databases of language cognition and use effective multi-lingual computational tools for estimating word probabilities.
 Fortunately, such datasets and computational tools are available. For instance,  existing multi-lingual LLMs % (\texttt{bert-base-multilingual-cased (m-BERT)}, \citet{devlin2018bert}; \texttt{xlm-roberta-large (XLM)}, \citet{conneau2020unsupervised})
  are highly capable of processing and understanding text in multiple languages, further helping compute sentence-level metrics we propose. %It is trained on a large dataset of text in multiple languages and it is able to identify the language of the input text and process it accordingly. The architecture of a multi-lingual transformer model is similar to that of a standard transformer model. 
 %Multi-lingual transformer models have some additional features that allow them to handle multiple languages. 
 %Either \texttt{m-BERT} or \texttt{XLM} is expected to compute sentence-level probability and similarity precisely for a range of languages. 
 There are also multilingual databases on language comprehension and processing. For instance, the Multilingual Eye-movement Corpus (MECO) is a collection of eye-tracking data that has been collected from participants reading texts in 13 languages \citep{siegelman2022expanding}. %This corpus provides a large, diverse set of data that can be used to study reading behaviour across languages. The MECO offers comprehensive data on the duration of fixations during sentence reading, applicable to multiple languages. 
 This is an ideal testing dataset for our research. %Multilingual fMRI dataset and other EEG datasets (Schoffelen et al. 2019; Bhattasali et al. 2020;  Li et al. 2022) can help in studying how the brain processes language and how different languages are represented in the brain. These existing databases on eye-movements and neural signals enable us to conduct experiments in which contextual computational measures are tested to show their predictability for the eye-movements or neural signals in multiple languages.
 %Our pilot study in English shows that both metrics are predictive of reading durations for processing sentences using the eye-movement data from the MECO \citep{siegelman2022expanding}.
The present study aims to compute sentence-level computational metrics and apply them to predicting sentence reading speed. By employing these metrics, we expect to obtain more accurate predictions of how sentences are processed and comprehended holistically. To evaluate the generalizability of our approach to various languages, we will undertake a multi-lingual investigation.%. To test the generalizability of our approach to different languages, we will conduct a multi-lingual investigation.
%The current study aims to extend the attenion-aware approach and develop enhanced measures to better predict how humans comprehend sentences as a whole. The proposed computational measures are expected to more accurately predict how words are understood and processed within a specific context. To evaluate the \textbf{generalization} of these measures across different languages, we will conduct a multi-lingual investigation. %Table \ref{table:sem} shows these measures and their algorithms.    

\section{Related Work}

Word surprisal, the negative logarithm of a word's probability in left context, is effectively used in testing expectation-based theories and predicting language processing \citep{hale2016information}. This metric signifies that predictable words require less cognitive effort to process. However, it remains an open question: which context aspects - lexical, syntactic, semantic, or conceptual - facilitate prediction or integration. While word surprisal accounts for the influence of the left context, this context is confined to a single sentence. Word surprisal has primarily been used to predict individual word processing times, such as the duration of reading a word. However, it falls short in predicting sentence-level measures, such as overall reading speed. Therefore, our goal is to develop metrics that go beyond within-sentence analysis to accurately predict reading speed. %While few models consider these context aspects and scopes, it is evident that the context surrounding a linguistic unit impacts its processing. %This suggests that a surprisal algorithm incorporating contextual information fully can better generalize unseen situations. Hence, the proposed ``attention-aware'' surprisal algorithm integrates contextual information, aligning with the attention mechanism in transformers and cognitive science.
As the language models developed so quickly, the methods for calculating word probabilities have been revolutionized in the transition from n-gram to Transformer \citep{wolf2020transformers}. %One common method uses n-gram models, which calculate the probability of a word based on the n-1 preceding words. However, there are other LMs that can estimate the probability distribution over successor words. 
For instance, recently deep-learning models have dominated NLP tasks and LLMs have been used as models for calculating the processing difficulty (\citealp{goodkind2018predictive}; \citealp{wilcox2020predictive}; \citealp{schrimpf2021neural}). The recent LLMs are effective to compute the probabilities of the next sentence in context. Moreover, ``surprisal'' is characterized as the negative logarithm of the probability of an event, and surprisal can be applied across different linguistic levels, including sentences, discourse, or phonemes (\citealp{venhuizen2019expectation}; \citealp{pimentel2021surprisal}; \citealp{gwilliams2022extracting}). Provided we have sufficient computational resources, we can accurately calculate sentence-level surprisals for mutiple languages. % As new models of deep learning emerge, better models for computing next word probabilities and surprisal will also surely emerge. Similarly, we can apply the method of computing word-based surprisal in calculating sentence-level surprisal. .
 %Multilingual LLMs can help estimate sentence-level surprisal for mutiple languages.

Moreover, some effective methods have been proposed for computing semantic relevance among words within a sentence. The new approach involves integrating contextual information using the attention-aware method, resulting in more powerful contextual semantic relevance (\citealp{sun2023interpretable}; \citealp{sun2023attention}). However, we believe that the similar methods can be extended and modified to process entire sentences rather than merely words. By employing an ``attention-aware'' approach, we can calculate the semantic relevance among sentences, enabling us to measure the extent to which a target sentence is semantically connected to its neighboring sentences within the discourse. %Further, we can enhance this approach by integrating various sources of information based on attention-aware surprisal and semantic relevance metrics, creating a novel sentence-level metric. %The following discussion provides the rationale behind this comprehensive approach.

Reading difficulties stem from sources such as word-level factors, syntactic structure, and sentence-level elements. While context significantly affects sentence processing, effective computational metrics are needed to evaluate this. Sentence comprehension involves the mental processes when understanding language utterances, significantly influenced by context and other factors. Although studies have examined various aspects of comprehension such as word processing and syntactic integration (\citealp{carpenter1995language}; \citealp{kamide2008anticipatory}; \citealp{altmann2009incrementality} \citealp{rohde2002connectionist}; \citealp{hale2016information}; \citealp{gibson1998linguistic}; \citealp{mcrae2013constraint}), there is a dearth of computational methods to measure context impact and research on comprehending sentences as a whole. This study aims to address this gap by developing computational sentence-leve metrics for evaluating entire sentences across languages.

Specifically, entire sentence comprehension and processing may be influenced by factors like sentence expectations from previous context and memory integration based on semantic relatedness. Word-level surprisal and semantic relevance have been shown to independently affect processing words in context. Words that are both semantically relevant and surprising may create more complex processing situations, requiring more cognitive resources. This study aims to explore whether sentence-level surprisal and semantic relevance are interactive in processing sentences as a whole. Moreover, sesearch on computational approaches to cognitive sciences has predominantly focused on the English language, leading to a lack of investigation in other languages to test the generalizability of findings on surprisal prediction effectiveness \citep{blasi2022over}. %The success of computational models for predicting cognitive processes in English does not guarantee success in other languages, as English is unique and may not be representative of other languages. The linguistic habits of English speakers and the characteristics of the English language have consequences beyond the study of language, and may warp research programs and lead to over-generalizations.
 It is essential to test hypotheses in multiple languages to enable cross-lingual \textbf{generalizations} using computational models or metrics.% Moreover, while some studies have carried out cross-linguistic research using transformer pretrained language models, others have only used \texttt{correlation} to confirm predictive power of some given computational measures (\citealp{hollenstein2021cmcl}; \citealp{salicchi2021looking}). Some regression models should be chosen to do the task. An optimal regression model should include proper control predictors and random variables. Unfortunately, existing studies have struggled to secure optimal regression models when estimating computational measures' predictive power. 

\section{Materials and Methods}
\subsection{Testing datasets}
Multilingual databases facilitate cost-effective and reliable cognitive and neural language testing. The Multilingual Eye-tracking Corpus (MECO) is particularly useful due to its diverse range of 13 languages and eye-tracking data indicating cognitive effort in processing \citep{siegelman2022expanding}. Each language's participants read 12 encyclopedia entries of 2000 tokens, resulting in 36000 tokens in total. These entries were similarly complex across languages and generated about 70,000 to 80,000 eye-tracking data points per language. MECO was chosen for its language diversity, large native speaker counts, and ample eye-tracking data. \textbf{Reading speed (or rate)} has been extensively explored, typically employing text or individual sentence as the unit of interest (\citealp{miller1971measurement}; \citealp{carver1976word}; \citealp{biancarosa2005speed}; \citealp{brysbaert2019many}; \citealp{siegelman2022expanding}). 

This study focuses on \textbf{sentence reading speed} (= \textit{the number of words in this sentence / total fixation duration of a sentence}) - \textbf{word number per second (for a sentence or text)}.  When the reading speed is low, readers use more time to process this sentence. In contrast, a higher sentence reading speed indicates that less time is used to process every word for this sentence. While reading speed is not a direct oculomotor measure, it is influenced by oculomotor factors such as fixation duration, saccade durations, and the tendency to skip words. Reading speed is interconnected with these oculomotor measures, as they collectively contribute to how quickly and efficiently one can read and understand a sentence or text. Fixation duration for word merely represents the difficulty for an individual word. In contrast, the reading speed is considered as the \textbf{overall difficulty when readers comprehend a sentence/text as a whole}. %In this sense, reading speed is a broader measure of language comprehension and processing at the sentence or text level, distinct from the measure of fixation duration for word. 
While reading speed measures the number of words processed per minute or second across sentences, word reading duration assesses the processing difficulty of a single word. As these measures are fundamentally different, word surprisal predicts the processing of individual words and does not extend to sentence-level processing (More are seen in \textbf{Appendix A}). We need to figure out practical sentence-level metrics to estimate reading speed.  

\subsection{Computing sentence surprisal}

We employed two multilingual LLMs: \texttt{m-BERT} and \texttt{mGPT} to compute sentence surprsial. Sentence surprisal is the negative logarithm of next sentence probability ($-log(p(sentence | left \quad context))$), which is similar to word surprisal. \texttt{Multi-lingual BERT} (\texttt{m-BERT}) \citep{devlin2018bert} can be employed to compute word-level or sentence-level surprisal for various languages. BERT, a masked language model, has good performance in a number of NLP tasks (\citealp{salazar2019masked}; \citealp{kalyan2021ammus}), and can be used to estimate next word probabilities. We used the state-of-the-art BERT model (i.e., \texttt{multilingual-bert-uncased}) because it can be consistently applied in different languages in order to compute \texttt{word surprisal}. However, it seems a little difficult to use it to compute sentence probability immediately using \texttt{m-BERT}. Fortunately, we can take some strategies to allow \texttt{m-BERT} to approximate next sentence probabilities. For instance, we gauged the joint probability of an entire sentence conditioned on its preceding context. This is achieved by using the chain rule of probability, breaking the sentence into individual tokens, and computing the probability of each token given all preceding tokens (and the context). 

Moreover, GPT is essentially an autoregressive model based on the Transformer architecture, trained on a language modeling task, where the objective is to predict the next word in a sequence given the preceding words. The model learns to assign probabilities to words based on the context. %To compute the surprisal of a sentence, we first need to compute the probability of the sentence given the preceding context. We do this by feeding the words of the sentence one-by-one into GPT and multiplying the probabilities that GPT assigns to each word in terms of the chain rule. 
The method of chain rule can be introduced in applying \texttt{mGPT} to compute sentence surprisal \cite{shliazhko2022mgpt}. The only difference is that GPT employs the language modeling to compute word probability, but BERT uses masked language model to calculate word probability. 

To compute sentence surprisal, we also proposed the other methods to compute sentence surprisal. The three methods are summarized as follows: the first one is to use \textbf{chain rule (CR)}, which was mentioned above. Sentence surprisal is calculated by tokenizing a sentence into individual elements and computing the joint probability of the sentence given a context. This is done by multiplying the conditional probabilities of each token given its preceding tokens and the context, applying the chain rule of probability to language sequences. The method can be computed based on either \texttt{m-BERT} or \texttt{mGPT}. The second method is to use ``next sentence prediction'' (\textbf{NSP}) mechanism in \texttt{m-BERT} to compute sentence surprsial. Third, sentence surprisal is also quantified by computing the \textbf{ Negative Log-Likelihood (NLL)} of the sentence when conditioned on its context. NLL is used as a loss function to measure
how well the model's predictions align with the actual data. For a given
sequence of tokens, NLL is a measure of how surprised the model is by the
actual sequence. The details on the three methods are seen in \textbf{Appendices B \& C}. %This is done by subtracting the NLL of the context from the NLL of the context with the sentence, resulting in the conditional NLL, which is negated to determine surprisal.  

Overall, we employed three methods for computing sentence surprisal, utilizing two multilingual LLMs: \texttt{m-BERT} and \texttt{mGPT}. ``Sentence surprisal'' quantifies \textbf{the level of unpredictability or information associated with encountering a given sentence}. The CR and the NLL approach were applied to both LLMs for this purpose. In contrast, the NSP method was exclusively implemented on \texttt{m-BERT}. The methods and models are shown in the {\color{blue}{Panel A}} of Fig. \ref{fig:computation}.   %The sentence surprisal represents the expectation

\subsection{Computing attention-aware sentence relevance}

\begin{figure*}[htp]%[!h]%[htp]
	\vskip -0.06in
	\centering
	
	\includegraphics[width=0.82\textwidth]{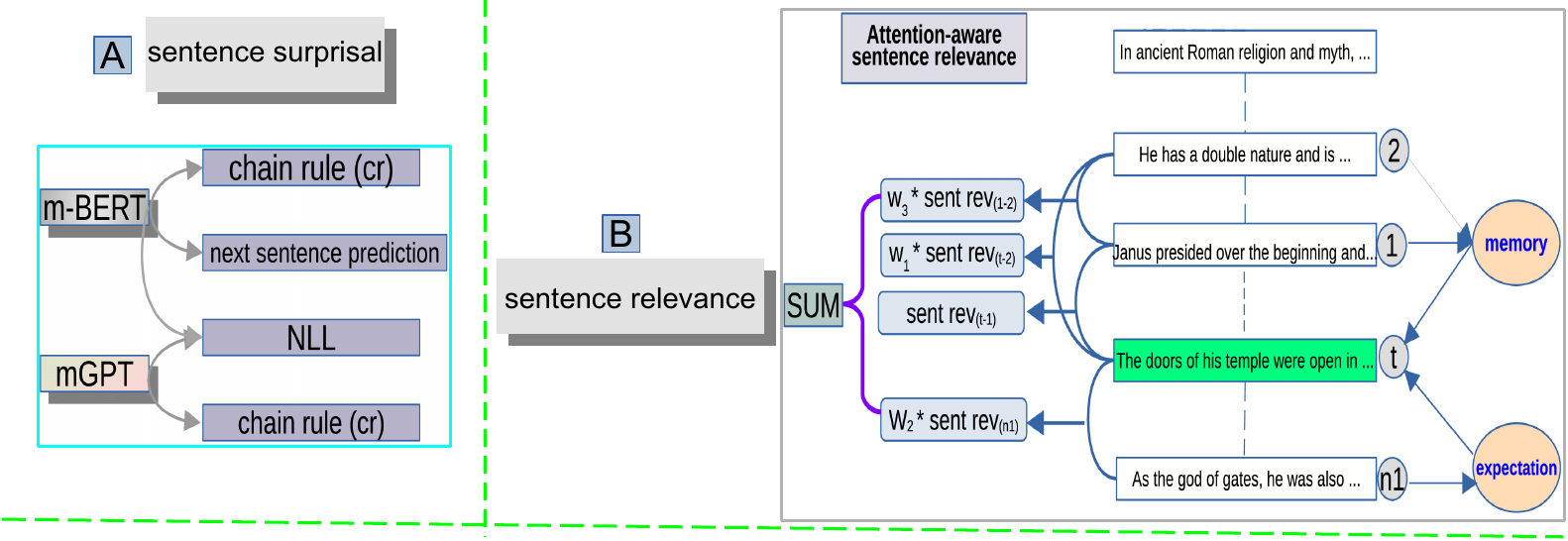}
	
	\caption{The computational methods in the present study}
	
	\label{fig:computation}
\vskip -0.11in	
\end{figure*}
%\vspace{-1.5mm}

This section elaborates on the computation of sentence semantic relevance. Consider a window comprising four sentences, as depicted in {\color{blue}{Panel B}} of Fig. \ref{fig:computation}, where the objective is to evaluate the semantic relationship of the target sentence with the adjacent three sentences within a window. The computation unfolds in two primary steps. Initially, embeddings for each sentence within the window are generated using either \texttt{m-BERT} or \texttt{mGPT}. Subsequently, the application of \texttt{cosine similarity} to the sentence embeddings derived from \texttt{m-BERT} or \texttt{mGPT} serves as a conventional approach to ascertain the semantic similarity between any pair of sentences, as detailed in \textbf{Appendix D}. Nonetheless, this step is to obtain the relevance of two sentences in this window. Our objective, however, is to compute the semantic relatedness between a target sentence and its contextual sentences, which is another sentence-level metric. 

Upon calculating the semantic similarity values for any pair of sentences within this window, we employ an ``attention-aware'' approach to process the four similarity values. The ``attention-aware'' approach has been successfully applied in computing word-level metrics for predicting reading (\citealp{sun2023interpretable}; \citealp{sun2023attention}). %Regarding sentence relevance, after obtaining a value of similarity between a target sentence and its left sentence, we can incorporate the semantic relevance from its contextual sentences. 
Specifically, we look at the preceding two sentences and the following sentence. %The similarity score for any two sentences is computed by taking the dot product between the first embedding of the concatenated input and the second embedding of the concatenated input.
Subsequently, we obtain four cosine similarity values for these sentences. Each value is then weighted differently based on its distance from the target sentence. It is formalized as (1).
$$atten\_sentrev = \text{similarity}_{(s,c)} \cdot W_{(s,c)}\qquad (1)$$ 
Here ``W'' represents weights according to the positional distance between the target sentence (``s'') and its contextual sentences (``c''). The range of weights is between 1/2 and 1/3 (i.e the neighboring sentence is multiplied by 1/2, but the non-neighboring one is by 1/3.). As the distance from the target sentence is large, the weight is given with smaller one. The weights used for attention-aware metrics mimic human forgetting mechanism, a representation of memory retention decline over time (e.g., the mean decline rate is 1/3 for three days), where retained information halves after each day within a span of several days \citep{loftus1985evaluating}, shown in Fig. \ref{fig:memory} (see \textbf{Appendix D}). %Different weights actually consider the sentence order and its semantic relevance with the target one. 
Unlike the attention weights in transformers, which are computed via neural networks and are not easily interpretable. %These weights are notably more interpretable, especially from cognitive and linguistic viewpoints. The rationale behind this interpretability is the observed decrease in influence a linguistic unit has on the target unit as their distance increases. 
Each similarity value is multiplied by its corresponding weight, and the results are then aggregated to produce a single value. This value signifies \textbf{the semantic relevance of the target sentence within its context}(i.e.,  ``sentence relevance''). Leveraging various sentence embeddings created by \texttt{m-BERT} and \texttt{mGPT}, we are able to derive two distinct metrics of ``sentence relevance'' for the identical target sentence within its specific context. The computation is shown as {\color{blue}{Panel B}} of Fig. \ref{fig:computation}. %More details on sentence similarity is seen in \textbf{Appendix B}.

The method proposed for computing attention-aware metrics was not invloved in introducing attention layers in Transformer. The term ``attention-aware'' sounds similar to attention because the method works as well as attention in incorporating contextual information. The ``attention-aware'' approach can work as a memory agent (see \textbf{Appendix D}). This approach considers both preceding and following sentences as potential sources of contextual information. The relationship between these contextual sentences and the target sentence is then weighted based on their positional distance and aggregated to create a comprehensive measure of semantic relevance within the discourse. This approach is highly \textbf{explainable} from linguistic and cognitive perspectives. Table \ref{table:sem} provides an overview of the metrics in our analysis.

\begin{table}[ht]
	\centering
	\caption{The models and statistical analysis}
	\label{table:sem}
	\resizebox{0.58\textwidth}{!}{%
		\begin{tabular}{||c c c c||} 
			\hline
			\textbf{Method} & \textbf{Measure} &  \textbf{Equation} & \textbf{Statistical Analysis} \\
			\hline\hline
			sentence surprisal & \makecell{chain rule (CR), \\ next sentence prediction (NSP),\\ negative loglikelihood (NLL)} &  $-\log_{2}p(\text{sentence} \,|\, \text{left context})$  & GAMM \\
			\hline
			attention-aware & sentence semantic relevance & $\sum{C_{(s, c)} \cdot W_{(s, c)}}$ & GAMM \\
			\hline\hline
		\end{tabular}%
	}
	\vskip -0.09in
\end{table}

\subsection{Statistical method and model comparison}
We employed Generalized Additive Mixed Models (GAMMs) \citep{wood2017generalized} to investigate the predictive power of sentence reading speed. The present study included these control predictors (mean word length/frequency for a sentence) and random variable (e.g.,``languages'', ``participants'' ).  Including these variables is crucial for achieving optimal GAMM fitting. %The past studies may not have reached optimal fitting due to the absence of either random variables or control predictors (\citealp{wilcox2020predictive}; \citet{oh2023does}).

The performance of GAMM models was evaluated using difference in \texttt{AIC} (Akaike's Information Criterion) (i.e.,\texttt{$\Delta$}AIC
) between a base GAMM and a full GAMM served as a measure to assess the effectiveness of a metric. A smaller \texttt{$\Delta$}AIC value indicates that the measure we are interested in estimates provide more accurate predictions of the response variable compared to the baseline model. The difference in AIC between a base GAMM and a full GAMM served as a measure to assess the effectiveness of a computational measure. A smaller (negative) \texttt{$\Delta$}AIC value indicates that the measure we are interested in estimates provide more accurate predictions of fixation duration compared to the baseline model. The more details on statistical methods can be seen in \textbf{Appendix E}. %The regression models used in the relevant studies (\citealp{wilcox2020predictive}; \citet{oh2023does}) may not be optimal due to the absence of random variables or control predictors. Additionally, the evaluation criteria (log-likelihood) employed in these studies may have certain limitations. %The regression models in the relevant studies (\citealp{wilcox2020predictive}; \citet{oh2023does}) may not be optimal because of lack of random variables or control predictor, and the evaulation criteria (log-likelihood) is also problematic. 

\section{Results}

\subsection{Overall performance}

We fitted eight GAMM models to analyze these sentence-level metrics as the main predictors of the response variable (\textit{sentence reading speed}). The GAMM models also include \textit{mean word length} and \textit{mean word frequency} \footnote{\scriptsize{``mean word length''  refers to the average length of all words within a sentence (i.e., \textit{the sum of length of all words / word number}), while ``mean word frequency'' represents the average of the sum of normalized frequencies for all words within a sentence (i.e., \textit{the sum of normalized word frequencies / word number}). The word frequency information across languages is obtained from \url{https://opus.nlpl.eu/index.php}.\citep{lison2016opensubtitles2016}}}, and \textit{participant} as a random effect. This is what an optimal GAMM formula looks like: \textit{reading\_speed $\sim$ s (mean\_word\_length), s(mean\_log\_wordfreq) + s(computational metric, language, by = languages) + languages + s (participant, bs=``re''), data=data} (here, \texttt{s} = smooth; \texttt{`by = languages'}, 
it facilitates the inclusion of responses at each language level in the model as distinct terms. Concurrently, smooths by language are employed to capture and model the nuanced variations around these language-specific responses; \texttt{re} = random effect, random slope adjusting the slope of the trend of a numeric predictor). The base GAMM fitting is depicted as: \textit{reading\_speed $\sim$ s (mean\_word\_length), s(mean\_log\_wordfreq) + languages + s(participant, bs=``re'')}. The \texttt{$\Delta$}AIC represents the difference in \texttt{AIC} values between a full model and a base model. % The other GAMM fitting appears as: \textsl{reading\_rate $\sim$ s (mean\_word\_length), s(mean\_log\_wordfreq) + te(sentence surprisal, sentence relevance) + s (participant, bs=``re''), data=data} (here, \texttt{te} = tensor interaction). 

First, we examine whether a variable is significant or not. When a variable in GAMM is significant, its \textit{p}-value is smaller than 0.01. These GAMM fittings show that the control predictors, namely \textit{mean word length} and \textit{mean word frequency}, are consistently significant across all cases. This indicates that the comprehension of sentences in the context of naturalistic discourse reading is significantly shaped by both the average length and frequency of the words employed. Specifically, the \textit{mean word length} exerts a negative influence on the speed of sentence reading; in essence, sentences composed of longer words are read more slowly. Conversely, the \textit{mean word frequency} positively impacts reading speed, meaning that sentences containing words that are more frequently used facilitate faster reading. Therefore, when the average frequency of words in a sentence is high, it enables readers to process the text more swiftly. The effects of word length and frequency are particularly notable in word processing during naturalistic discourse reading. As such, sentence processing and word processing share a great deal of similarity. GAMM fitting results also show that the majority of the metrics we proposed to compute by \texttt{m-BERT} or \texttt{mGPT} are capable of predicting the reading speed data quite well. The random effect of the \textit{participant} is significant in all GAMM models.  %Unfortunately, the prediction performance of \texttt{XLM} falls short when it comes to calculating sentence surprisal and relevance. The primary reason for this shortcoming seems to lie in the notably low variance of the data on sentence probability and similarity, as computed by \texttt{XLM}, which means that \texttt{XLM} seems not be good at computing sentence probability or sentence similarity. 
%In comparison, the chain rule (CR) method of \texttt{m-BERT} notably outperforms in predictive metrics, in comparison with two methods in \texttt{mGPT}. While \texttt{NSP} in BERT also demonstrates significant predictive performance, it clearly trails behind \texttt{m-BERT} or \texttt{mGPT} in this regard. However, when it comes to sentence similarity, \texttt{m-BERT} remarkably outdoes \texttt{mGPT}. Other refined measurements exhibited marginal enhancements in comparison to the baseline surprisal or relevance metrics. %For a clearer comparison, we will use sentence surprisal as the baseline. As such, we did not include the results from \texttt{MiniLM}'s sentence surprisal in our GAMMM fitting. Additionally, metrics derived from \texttt{XLM} will also be excluded.

 %The results indicate that sentence surprisal tends to have a negative effect on sentence reading speed across languages. By contrast, sentence relevance seems to have a positive effect on sentence reading speed for mutiple languages. %This contour plot demonstrates that   

We further compared the performance of the GAMM fittings with different metrics. Table \ref{tab:aic} presents the results on \texttt{$\Delta$}AIC for comparing these GAMM fittings. %\texttt{AIC} measures the balance between model complexity and goodness-of-fit and can be used to choose the best-fitting model. 
Lower \texttt{$\Delta$}AIC value indicates a better GAMM fitting. It also suggests that the computational metric performs better than others. %\texttt{AIC} is often preferred over \texttt{log-likelihood} for model comparison because it takes into account both goodness of fit and model complexity. \texttt{AIC} is also preferred over \texttt{BIC} because the latter penalizes models more heavily for additional parameters.
The basis for comparison is the consistent data point numbers (n = 34571) and identical elements in each GAMM fitting. %The $\Delta$AIC values for each GAMM model were computed by subtracting the AIC of the base GAMM model from that of a full GAMM model. The base GAMM model does not incorporate the variables of our interest. A smaller $\Delta$ AIC indicates better performance. 
As illustrated in Table \ref{tab:aic}, the result reveals that sentence surprisal, as calculated using three distinct methodologies — namely, CR, NLL, and NSP — proves to be effective in predicting reading speed. Among the evaluated metrics of sentence surprisal, the performance of \texttt{m-BERT}, when applying the chain rule, stands out as the most effective. Conversely, sentence surprisal calculations using \texttt{mGPT} with NLL were found to be ineffective in predicting reading speed. We therefore skiped the result of mGPT with NLL. The computation of sentence relevance, whether through \texttt{m-BERT} or \texttt{mGPT}, demonstrates viability in prediction accuracy. Notably, the integration of both sentence surprisal and sentence relevance into the GAMM significantly enhances predictive performance, surpassing the results achieved when these metrics are applied independently. %we found that the sentence surprisal computed by the three methods (i.e., chain rule, NLL and NSP) could work in predicting reading rate. We found that sentence surprisal computed by \texttt{m-BERT} with chain rule performs best among the sentence surprisal metrics. Note that sentence surprisal computed by \texttt{mGPT} with NLL did not work. Sentence relevance computed either by \texttt{m-BERT} or \texttt{mGPT} can work in prediction. When sentence surprisal and sentence relevance are put togather in the GAMM model, it work much better than individually. %, specifically those incorporating sentence surprisal or sentence relevance, outperformed the metrics that did not take into account attention-aware weights. 

\begin{table}
	\centering
	\vskip -0.06in
	
	\caption{\texttt{$\Delta$}AIC for GAMM fittings with different computational measures on the MECO  (n = 34571) (CR = Chain Rule; NLL = Negative Log-Likelihood; NSP = Next Sentence Prediction)}
	\scalebox{0.76}{
	\begin{tabular}{||c c||}
		\hline
		GAMM fittings (sentence reading speed)       &  \texttt{$\Delta$}AIC

  \\
		\hline \hline
		%\topmidheader{5}{\textcolor{brown}{[\texttt{AIC}]}}
		\textcolor{green}{sentence surprisal (m-BERT-CR)}    & \textcolor{green}{-670.06}    \\
		sentence surprisal (m-BERT-NSP)     & -119.94     \\
		sentence surprisal (m-BERT-NLL) &  -506.85\\
		sentence surprisal (mGPT-CR)    & -393.56    \\
	%	sentence surprisal (mGPT-NLL)     & NA     \\
		\hline
		attention-aware sentence relevance (m-BERT)        & -445.86       \\
	    attention-aware sentence relevance (mGPT)        &  -294.1   \\
	    \hline
	\textcolor{blue}{\makecell{sentence surprisal (m-BERT-CR), \\sentence relevance (m-BERT)}}       & \textcolor{blue}{-1397.6}    \\
	%	enhanced sentence surprisal (m-BERT-NSP)       & -159.9     \\
	%	enhanced sentence surprisal (mGPT-CR)       & -498.3     \\
	%	enhanced sentence surprisal (mGPT-NLL)       & -598.2     \\
		%\hline
		%	\texttt{te} (sentence surprisal, sentence relevance) (m-BERT-CR)      & -983.3    \\
		%	\texttt{te} (sentence surprisal, sentence relevance) (mGPT-CR)      & -195.3    \\
		\hline \hline
		%	\midheader{5}{III-V Compounds}
		%	indium phosphide& InP    & 1.27 & d \\
		%	indium arsenide & InAs   & 0.355& d \\
		%	gallium nitride & GaN    & 3.37 & d \\
		%	gallium arsenide& GaAs   & 1.42 & d \\
		%	aluminium nitride & AlN  & 6.2  & d \\
	\end{tabular}}
	\label{tab:aic}
%	\addtabletext{``ASR\_W'' = Attention-aware semantic relevance (weak); ``ASR\_L'' = Attention-aware semantic relevance (large weights); ``ASR\_S'' = Attention-aware semantic relevance (small weights)}
\vskip -0.11in
\end{table}

 \begin{figure*}[htp]%[htp]
 	%\vskip 0.16in
 	\centering
 	
 	\includegraphics[width=1.02\textwidth]{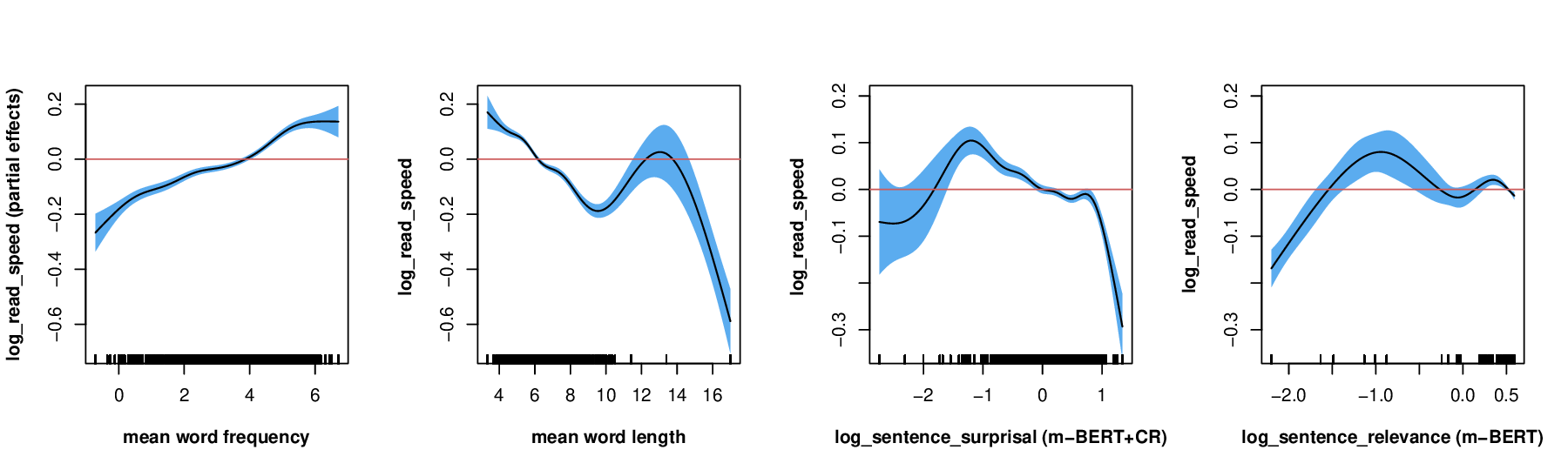}
 	
 	\caption{The overall partial effects of the primary predictors—sentence surprisal and sentence relevance—on reading speed across languages. The \textit{x-axis} denotes the metric, and the \textit{y-axis} represents the reading speed. Sentence surprisal and sentence relevance are transformed by logarithm in order to be get closer normal distribution, further having better fittings. Each curve depicts the relationship between a predictor variable and the response variable, reading speed. Steeper slopes on these curves indicate a stronger relationship between the predictor and reading speed, while flatter slopes suggest a weaker effect.}
 	
 	\label{fig:whole}
 	\vskip -0.06in
 \end{figure*}

The overall partial effects of the metrics of our interest (excluding the ``languages'' factor) in GAMM fittings are shown in Fig. \ref{fig:whole}. The results reveal that sentence surprisal (by \texttt{m-BERT} and CR basically adversely affects reading speed across 13 languages, indicating that a higher level of surprisal for a sentence is associated with slower/lower reading speed. On the other hand, sentence relevance (by \texttt{m-BERT}) positively influences reading speed across these languages, suggesting that greater relevance for a sentence facilitates higher/faster reading speed.

\subsection{Performance in individual languages}

The results from fitting the GAMM offer valuable insights into the importance of the variables considered. %Using the \texttt{$\Delta$}AIC, we can determine the relative performance of different GAMM configurations. 
%We identified the optimal metrics for our study: sentence surprisal calculated using \texttt{m-BERT} with the chain rule (CR), and sentence relevance determined by \texttt{m-BERT}. This analysis enables us to select optimal GAMM structures for evaluating how computational metrics predict sentence reading speed across 13 languages.
 This section focuses on the performance of these metrics in each language (among 13 languages) using the similar GAMM analysis. The optimal GAMM for each language is specified as follows:\textit{reading\_speed $\sim$ s (mean\_word\_length), s(mean\_log\_wordfreq) + s(sentence \quad surprisal) + s(sentence  \quad relevance) + s (participant, bs=``re''), data= language\_data}. To calculate the \texttt{$\Delta$}AIC for ``sentence relevance'', we excluded $s(sentence \quad surprisal)$ from the model, comparing the \texttt{AIC} of the complete model with that of the model absent ``sentence surprisal''. Conversely, to determine the \texttt{$\Delta$}AIC for sentence surprisal,$s(sentence \quad relevance)$ is removed, and the \texttt{AIC} of the full model is subtracted from the model without ``sentence relevance''. The significance of either ``sentence surprisal'' or ``sentence relevance'' is assessed using the \textit{p}-value ( at \textit{p}-value of 0.01) and the shape of the curve. These metrics were selected to represent, respectively, the concepts of sentence surprisal and sentence relevance in our examination of each language. The results are illustrated in Fig. \ref{fig:effects} in \textbf{Appendix F}. It turns out that sentence surprisal and sentence relevance shows significance in the majority of languages.
%\vspace{-0.6cm}
%\vspace{-1.1pt}
 Moreover, sentence surprisal and sentence relevance are totally distinct metrics, and their overall correlation is -0.054, their correlation in each language is also remarkably small. This indicates that sentence surprisal and sentence relevance are distinct metrics, and more are seen in \textbf{Appendix G}.

\section{Discussion}

%\subsection{Predictability of sentence surprisal and relevance}

%Generally, the metrics computed by \texttt{m-BERT} or \texttt{mGPT} are able to predict reading rate. However, \texttt{m-BERT} excels in computing sentence surprisal and similarity, whereas \texttt{mPGT} falls short. Despite this, the method of ``NSP'' in \texttt{BERT} demonstrated the less effectiveness in prediction. The issue stems from 'NSP's design, which limits context consideration to the sentence immediately preceding the target. This narrow scope falls short of the expansive context humans naturally utilize when reading and interpreting text in real-world situations.

%However, sentence similarity derived from BERT models is not as effective as their sentence surprisal calculations. This is primarily because computing sentence or textual similarity is more complex than computing word or phrase similarity, making it a  little challenging for current pretrained language models to estimate sentence similarity accurately. Due to the lack of precise sentence similarity data, its predictability is not as significant as sentence surprisal in prediciting and explaining sentence reading durations. 

Word surprisal has been found to predict language comprehension across various datasets \citep{ryskin2023prediction}. Sentence prediction has also been found in impacting language comprehension(\citealp{goldstein2022shared}; \citealp{kriegeskorte2008representational}). However, sentence-level computational metrics have not been proposed previously. Our study found that the surprisal of a sentence can predict and clarify the difficulties readers encounter in comprehending a sentence as a whole. Specifically, if a sentence is highly predictable given its left context, readers may feel easier to comprehend or process, leading to higher/faster reading speed. On the other hand, if a sentence is more unpredictable (i.e., lower sentence surprisal), readers may be more likely to experience processing difficulty. %, as they try to reconcile the new information with their existing mental representation of the text. 
This can lead to slower/lower reading speed and potentially disrupt the flow of reading. The ability to predict the next sentence may be a fundamental aspect of language comprehension, as it allows users to anticipate upcoming information, and to integrate it smoothly with the preceding information. The impact of sentence surprisal on reading aligns closely with the influence of word surprisal on the reading process(\citealp{goodkind2018predictive}; \citealp{gotlieb2023testing}). Prediction occurs not only at the level of individual words, but also at the level of entire sentences in discourse. Prediction could be a key mechanism underlying human language comprehension, which makes predictions about the likely content of the upcoming input. %, and adjusts those predictions as new information is encountered.

Numerous studies have also identified the influence of context on sentence processing (\citealp{cohen1992context}; \citealp{grisoni2017neural}). However, we are the first to propose to compute sentence-level semantic relevance. The sentence relevance computed by the ``attention-aware'' method can effectively help interpret and predict the processing difficulties readers face when comprehending sentences in their entirety. When a sentence is more semantically relevant to the context (in discourse) in which it appears, it is more likely to be processed quickly and accurately, and readers are more likely to understand its meaning without having to invest a lot of cognitive effort (i.e. higher reading speed). Conversely, when a sentence is irrelevant with the context (i.e., lower sentence relevance), it may slow down reading speed and require more cognitive effort. The impact of sentence relevance on reading closely resembles the effect of word semantic relevance on the reading (\citealp{sun2023attention}; \citealp{sun2023interpretable}). There are several possible reasons why relevant sentences tend to be processed more quickly. According to the research of discourse structure, relevant sentences can easily create discourse coherence. One key factor is the activation of mental representations related to the topic or theme of the text \citep{traxler2011introduction}. When a sentence is relevant to the context, it could activate mental representations through memorizing the context, making it easier for readers to process the sentence. This activation can lead to faster reading speed, as readers are able to effectively activate the memory of the context and quickly integrate it into their mental representation of the text. In contrast, irrelevant sentences may require more cognitive effort to process, as they could not activate the memory from the context and do not fit smoothly into the mental representation of the discourse that readers are constructing. This processing difficulty can slow down reading speed.%, potentially disrupting the flow of reading and making it more difficult to process.

%Additionally, relevant sentences are usually more predictable than irrelevant ones, correlating semantic relevance with sentence surprisal. This predictability, derived from context consistency, accelerates reading speed and enhances comprehension. %relevant sentences tend to be more predictable than irrelevant sentences. In other words, semantic relevence of sentence is highly correlated with sentence surprisal. Predictability is a key factor that influences reading speed, as readers are able to process predictable words and sentences more quickly than unpredictable ones. When a sentence is semanticlly relevant to the context, it is more likely to be predictable, as it follows the patterns and themes established by the preceding sentences. This predictability can further enhance reading speed and comprehension. %In summary, the contextual sentence semantic relevance is an important factor that influences how readers comprehend sentences as a whole. When a sentence is relevant to the context and is consistent with expectations set by previous sentences, it is more likely to be processed quickly and accurately, leading to a better understanding of the overall text.

%\subsection{Enhanced metrics}

The expectation and memory could be mutually interacted \citep{ryskin2023prediction} in language comprehension/processing.  By combining sentence surprisal and sentence relevance (shown in Table \ref{tab:aic}),  it can encompass both processing difficulty for an entire sentence, and could better simulate human language processing. %When combining sentence surprisal and contextual sentence relevance to create enhanced metric, two important metrics are captured: the predictability of the sentence and its contextual fit within the discourse. 

\section{Conclusion}
This study presented two sentence-level metrics for predicting human comprehension of sentences as a whole. The methods of computing sentence surprisal worked well, and the attention-aware method allowed for computing contextual information for sentence-level semantic relevance. The results show that both sentence surprisal and sentence relevance were highly capable of predicting human sentence reading speed. All of these methods also exhibited strong generalization capabilities across languages. The findings showed that these sentence-level metrics are informative features for modeling entire sentence comprehension by humans across languages. Our work highlighted the potential of combining computational models with cognitive models to better explain and predict human language comprehension/processing, and further to develop more effective NLP and AGI systems.

\section{Limitations}

The study introduces innovative approaches to predicting entire sentence comprehension using computational sentence-level metrics. However, it comes with several limitations. Primarily, the methodology depends on specific multilingual LLM (\texttt{m-BERT} and \texttt{mGPT}). The variation in proficiency across different languages can result in disparate abilities to process these languages. In essence, sentence surprisal or sentence relevance might not be estimated with equal precision across all languages using multilingual LLMs. This limitation could affect the applicability of our metrics across varied linguistic structures and cultural contexts, leading to varying performance of our metrics in different languages. Moreover, the attention-aware method, while effectively capturing some aspects of contextual information, may oversimplify the complex dynamics of human cognitive mechanisms on language comprehension and processing. %The interpretability and generalization of the proposed metrics, although promising, require further validation across a wider array of languages and text types to ascertain their universal applicability. %Additionally, the study's focus on sentence-level processing might overlook the nuances of word-level interactions and their cumulative effect on sentence comprehension. 
Addressing these limitations could pave the way for more comprehensive models that better mirror the intricacies of human language comprehension and processing.
 
\section*{Data Availability}
The code and data in this study is available at: {\footnotesize{\url{https://github.com/fivehills/Sentence-relevance-and-sentence-surprisal/tree/main}}}.

%This study was supported by \textit{China Postdoctoral Science Foundation} (Outstanding Fund No. \texttt{2018T110581}) We appreciate helps provided by Prof. Haitao Liu at Zhejiang University, China. Many thanks go to Ms. Yiyun Zhou, and Ms. Mengjie Xu, and Dr. Yi Shan for their helps in text processing and annotations. 

%\bibliographystyle{acl_natbib}
\bibliography{reference.bib}

\begin{thebibliography}{70}
\expandafter\ifx\csname natexlab\endcsname\relax\def\natexlab#1{#1}\fi

\bibitem[{Altmann and Mirkovi{\'c}(2009)}]{altmann2009incrementality}
Gerry~TM Altmann and Jelena Mirkovi{\'c}. 2009.
\newblock Incrementality and prediction in human sentence processing.
\newblock \emph{Cognitive Science}, 33(4):583--609.

\bibitem[{Arehalli et~al.(2022)Arehalli, Dillon, and
  Linzen}]{arehalli2022syntactic}
Suhas Arehalli, Brian Dillon, and Tal Linzen. 2022.
\newblock Syntactic surprisal from neural models predicts, but underestimates,
  human processing difficulty from syntactic ambiguities.
\newblock \emph{arXiv preprint arXiv:2210.12187}.

\bibitem[{Baayen and Linke(2020)}]{baayen2020introduction}
R~Harald Baayen and Maja Linke. 2020.
\newblock An introduction to the generalized additive model.
\newblock \emph{A Practical Handbook of Corpus Linguistics}, pages 563--591.

\bibitem[{Baddeley(2010)}]{baddeley2010working}
Alan Baddeley. 2010.
\newblock Working memory.
\newblock \emph{Current Biology}, 20(4):136--140.

\bibitem[{Biancarosa(2005)}]{biancarosa2005speed}
Gina Biancarosa. 2005.
\newblock Speed and time, texts and sentences: Choosing the best metric for
  relating reading rate to comprehension.
\newblock \emph{Written Language \& Literacy}, 8(2):3--24.

\bibitem[{Blasi et~al.(2022)Blasi, Henrich, Adamou, Kemmerer, and
  Majid}]{blasi2022over}
Dami{\'a}n~E Blasi, Joseph Henrich, Evangelia Adamou, David Kemmerer, and Asifa
  Majid. 2022.
\newblock Over-reliance on english hinders cognitive science.
\newblock \emph{Trends in Cognitive Sciences}.

\bibitem[{Broderick et~al.(2018)Broderick, Anderson, Di~Liberto, Crosse, and
  Lalor}]{broderick2018electrophysiological}
Michael~P Broderick, Andrew~J Anderson, Giovanni~M Di~Liberto, Michael~J
  Crosse, and Edmund~C Lalor. 2018.
\newblock Electrophysiological correlates of semantic dissimilarity reflect the
  comprehension of natural, narrative speech.
\newblock \emph{Current Biology}, 28(5):803--809.

\bibitem[{Brysbaert(2019)}]{brysbaert2019many}
Marc Brysbaert. 2019.
\newblock How many words do we read per minute? a review and meta-analysis of
  reading rate.
\newblock \emph{Journal of Memory and Language}, 109:104047.

\bibitem[{Carpenter et~al.(1995)Carpenter, Miyake, and
  Just}]{carpenter1995language}
Patricia~A Carpenter, Akira Miyake, and Marcel~Adam Just. 1995.
\newblock Language comprehension: Sentence and discourse processing.
\newblock \emph{Annual Review of Psychology}, 46(1):91--120.

\bibitem[{Carver(1976)}]{carver1976word}
Ronald~P Carver. 1976.
\newblock Word length, prose difficulty, and reading rate.
\newblock \emph{Journal of Reading Behavior}, 8(2):193--203.

\bibitem[{Carver(1990)}]{carver1990reading}
Ronald~P Carver. 1990.
\newblock \emph{Reading rate: A review of research and theory.}
\newblock Academic Press.

\bibitem[{Cohen and Servan-Schreiber(1992)}]{cohen1992context}
Jonathan~D Cohen and David Servan-Schreiber. 1992.
\newblock Context, cortex, and dopamine: a connectionist approach to behavior
  and biology in schizophrenia.
\newblock \emph{Psychological Review}, 99(1):45.

\bibitem[{Crocker et~al.(2016)Crocker, Demberg, and
  Teich}]{crocker2016information}
Matthew~W Crocker, Vera Demberg, and Elke Teich. 2016.
\newblock Information density and linguistic encoding (ideal).
\newblock \emph{KI-K{\"u}nstliche Intelligenz}, 30:77--81.

\bibitem[{De~Varda and Marelli(2023)}]{de2023scaling}
Andrea De~Varda and Marco Marelli. 2023.
\newblock Scaling in cognitive modelling: A multilingual approach to human
  reading times.
\newblock In \emph{Proceedings of the 61st Annual Meeting of the Association
  for Computational Linguistics (Volume 2: Short Papers)}, pages 139--149.

\bibitem[{Demberg and Keller(2008)}]{demberg2008data}
Vera Demberg and Frank Keller. 2008.
\newblock Data from eye-tracking corpora as evidence for theories of syntactic
  processing complexity.
\newblock \emph{Cognition}, 109(2):193--210.

\bibitem[{Devlin et~al.(2018)Devlin, Chang, Lee, and
  Toutanova}]{devlin2018bert}
Jacob Devlin, Ming-Wei Chang, Kenton Lee, and Kristina Toutanova. 2018.
\newblock Bert: Pre-training of deep bidirectional transformers for language
  understanding.
\newblock \emph{arXiv preprint arXiv:1810.04805}.

\bibitem[{Gibson(1998)}]{gibson1998linguistic}
Edward Gibson. 1998.
\newblock Linguistic complexity: Locality of syntactic dependencies.
\newblock \emph{Cognition}, 68(1):1--76.

\bibitem[{Goldstein et~al.(2022)Goldstein, Zada, Buchnik, Schain, Price,
  Aubrey, Nastase, Feder, Emanuel, Cohen et~al.}]{goldstein2022shared}
Ariel Goldstein, Zaid Zada, Eliav Buchnik, Mariano Schain, Amy Price, Bobbi
  Aubrey, Samuel~A Nastase, Amir Feder, Dotan Emanuel, Alon Cohen, et~al. 2022.
\newblock Shared computational principles for language processing in humans and
  deep language models.
\newblock \emph{Nature neuroscience}, 25(3):369--380.

\bibitem[{Goodkind and Bicknell(2018)}]{goodkind2018predictive}
Adam Goodkind and Klinton Bicknell. 2018.
\newblock Predictive power of word surprisal for reading times is a linear
  function of language model quality.
\newblock In \emph{Proceedings of the 8th workshop on cognitive modeling and
  computational linguistics (CMCL 2018)}, pages 10--18.

\bibitem[{Gotlieb~Wilcox et~al.(2023)Gotlieb~Wilcox, Pimentel, Meister,
  Cotterell, and Levy}]{gotlieb2023testing}
Ethan Gotlieb~Wilcox, Tiago Pimentel, Clara Meister, Ryan Cotterell, and
  Roger~P Levy. 2023.
\newblock Testing the predictions of surprisal theory in 11 languages.
\newblock \emph{arXiv e-prints}, pages arXiv--2307.

\bibitem[{Grisoni et~al.(2017)Grisoni, Miller, and
  Pulverm{\"u}ller}]{grisoni2017neural}
Luigi Grisoni, Tally~McCormick Miller, and Friedemann Pulverm{\"u}ller. 2017.
\newblock Neural correlates of semantic prediction and resolution in sentence
  processing.
\newblock \emph{Journal of Neuroscience}, 37(18):4848--4858.

\bibitem[{Gwilliams and Davis(2022)}]{gwilliams2022extracting}
Laura Gwilliams and Matthew~H Davis. 2022.
\newblock Extracting language content from speech sounds: the information
  theoretic approach.
\newblock In \emph{Speech perception}, pages 113--139. Springer.

\bibitem[{Hale(2001)}]{hale2001probabilistic}
John Hale. 2001.
\newblock A probabilistic earley parser as a psycholinguistic model.
\newblock In \emph{The Second Meeting of the North American Chapter of the
  Association for Computational Linguistics}.

\bibitem[{Hale(2016)}]{hale2016information}
John Hale. 2016.
\newblock Information-theoretical complexity metrics.
\newblock \emph{Language and Linguistics Compass}, 10(9):397--412.

\bibitem[{Herman(1985)}]{herman1985effect}
Patricia~A Herman. 1985.
\newblock The effect of repeated readings on reading rate, speech pauses, and
  word recognition accuracy.
\newblock \emph{Reading research quarterly}, pages 553--565.

\bibitem[{Hohenstein and Kliegl(2014)}]{hohenstein2014semantic}
Sven Hohenstein and Reinhold Kliegl. 2014.
\newblock Semantic preview benefit during reading.
\newblock \emph{Journal of Experimental Psychology: Learning, Memory, and
  Cognition}, 40(1):166.

\bibitem[{Hollis and Westbury(2016)}]{hollis2016principals}
Geoff Hollis and Chris Westbury. 2016.
\newblock The principals of meaning: Extracting semantic dimensions from
  co-occurrence models of semantics.
\newblock \emph{Psychonomic Bulletin \& Review}, 23(6):1744--1756.

\bibitem[{Jackson and McClelland(1979)}]{jackson1979processing}
Mark~D Jackson and James~L McClelland. 1979.
\newblock Processing determinants of reading speed.
\newblock \emph{Journal of Experimental Psychology: General}, 108(2):151.

\bibitem[{Kalyan et~al.(2021)Kalyan, Rajasekharan, and
  Sangeetha}]{kalyan2021ammus}
Katikapalli~Subramanyam Kalyan, Ajit Rajasekharan, and Sivanesan Sangeetha.
  2021.
\newblock Ammus: A survey of transformer-based pretrained models in natural
  language processing.
\newblock \emph{arXiv preprint arXiv:2108.05542}.

\bibitem[{Kamide(2008)}]{kamide2008anticipatory}
Yuki Kamide. 2008.
\newblock Anticipatory processes in sentence processing.
\newblock \emph{Language and Linguistics Compass}, 2(4):647--670.

\bibitem[{Kennedy and Pynte(2005)}]{kennedy2005parafoveal}
Alan Kennedy and Jo{\"e}l Pynte. 2005.
\newblock Parafoveal-on-foveal effects in normal reading.
\newblock \emph{Vision Research}, 45(2):153--168.

\bibitem[{Kliegl et~al.(2007)Kliegl, Risse, and Laubrock}]{kliegl2007preview}
Reinhold Kliegl, Sarah Risse, and Jochen Laubrock. 2007.
\newblock Preview benefit and parafoveal-on-foveal effects from word n+ 2.
\newblock \emph{Journal of Experimental Psychology: Human Perception and
  Performance}, 33(5):1250.

\bibitem[{Kriegeskorte et~al.(2008)Kriegeskorte, Mur, and
  Bandettini}]{kriegeskorte2008representational}
Nikolaus Kriegeskorte, Marieke Mur, and Peter~A Bandettini. 2008.
\newblock Representational similarity analysis-connecting the branches of
  systems neuroscience.
\newblock \emph{Frontiers in systems neuroscience}, page~4.

\bibitem[{Levy(2008)}]{levy2008expectation}
Roger Levy. 2008.
\newblock Expectation-based syntactic comprehension.
\newblock \emph{Cognition}, 106(3):1126--1177.

\bibitem[{Lison and Tiedemann(2016)}]{lison2016opensubtitles2016}
Pierre Lison and J{\"o}rg Tiedemann. 2016.
\newblock Opensubtitles2016: Extracting large parallel corpora from movie and
  tv subtitles.

\bibitem[{Loftus(1985)}]{loftus1985evaluating}
Geoffrey~R Loftus. 1985.
\newblock Evaluating forgetting curves.
\newblock \emph{Journal of Experimental Psychology: Learning, Memory, and
  Cognition}, 11(2):397.

\bibitem[{McRae and Matsuki(2013)}]{mcrae2013constraint}
Ken McRae and Kazunaga Matsuki. 2013.
\newblock Constraint-based models of sentence processing.
\newblock In \emph{Sentence Processing}, pages 51--77. Psychology Press.

\bibitem[{Miller and Coleman(1971)}]{miller1971measurement}
Gerald~R Miller and Edmund~B Coleman. 1971.
\newblock The measurement of reading speed and the obligation to generalize to
  a population of reading materials.
\newblock \emph{Journal of Reading Behavior}, 4(3):48--56.

\bibitem[{Mitchell and Lapata(2010)}]{mitchell2010composition}
Jeff Mitchell and Mirella Lapata. 2010.
\newblock Composition in distributional models of semantics.
\newblock \emph{Cognitive Science}, 34(8):1388--1429.

\bibitem[{Oh and Schuler(2023)}]{oh2023does}
Byung-Doh Oh and William Schuler. 2023.
\newblock Why does surprisal from larger transformer-based language models
  provide a poorer fit to human reading times?
\newblock \emph{Transactions of the Association for Computational Linguistics},
  11:336--350.

\bibitem[{Pimentel et~al.(2021)Pimentel, Meister, Salesky, Teufel, Blasi, and
  Cotterell}]{pimentel2021surprisal}
Tiago Pimentel, Clara Meister, Elizabeth Salesky, Simone Teufel, Dami{\'a}n
  Blasi, and Ryan Cotterell. 2021.
\newblock A surprisal--duration trade-off across and within the world's
  languages.
\newblock \emph{arXiv preprint arXiv:2109.15000}.

\bibitem[{R~Anderson(1975)}]{r1975computer}
John R~Anderson. 1975.
\newblock Computer simulation of a language acquisition system: A first report.

\bibitem[{Rayner et~al.(2010)Rayner, Slattery, and
  B{\'e}langer}]{rayner2010eye}
Keith Rayner, Timothy~J Slattery, and Nathalie~N B{\'e}langer. 2010.
\newblock Eye movements, the perceptual span, and reading speed.
\newblock \emph{Psychonomic Bulletin \& Review}, 17(6):834--839.

\bibitem[{Rohde(2002)}]{rohde2002connectionist}
Douglas~LT Rohde. 2002.
\newblock \emph{A connectionist model of sentence comprehension and
  production}.
\newblock Carnegie Mellon University.

\bibitem[{Roland et~al.(2012)Roland, Yun, Koenig, and
  Mauner}]{roland2012semantic}
Douglas Roland, Hongoak Yun, Jean-Pierre Koenig, and Gail Mauner. 2012.
\newblock Semantic similarity, predictability, and models of sentence
  processing.
\newblock \emph{Cognition}, 122(3):267--279.

\bibitem[{Ryskin and Nieuwland(2023)}]{ryskin2023prediction}
Rachel Ryskin and Mante~S Nieuwland. 2023.
\newblock Prediction during language comprehension: what is next?
\newblock \emph{Trends in Cognitive Sciences}.

\bibitem[{Sakurai(2023)}]{Masato2023polychromy}
Masato Sakurai. 2023.
\newblock Parafovea.
\newblock In \emph{Encyclopedia of Color Science and Technology}, pages
  1319--1326. Springer.

\bibitem[{Salazar et~al.(2019)Salazar, Liang, Nguyen, and
  Kirchhoff}]{salazar2019masked}
Julian Salazar, Davis Liang, Toan~Q Nguyen, and Katrin Kirchhoff. 2019.
\newblock Masked language model scoring.
\newblock \emph{arXiv preprint arXiv:1910.14659}.

\bibitem[{Schotter et~al.(2012)Schotter, Angele, and
  Rayner}]{schotter2012parafoveal}
Elizabeth~R Schotter, Bernhard Angele, and Keith Rayner. 2012.
\newblock Parafoveal processing in reading.
\newblock \emph{Attention, Perception, \& Psychophysics}, 74(1):5--35.

\bibitem[{Schrimpf et~al.(2021)Schrimpf, Blank, Tuckute, Kauf, Hosseini,
  Kanwisher, Tenenbaum, and Fedorenko}]{schrimpf2021neural}
Martin Schrimpf, Idan~Asher Blank, Greta Tuckute, Carina Kauf, Eghbal~A
  Hosseini, Nancy Kanwisher, Joshua~B Tenenbaum, and Evelina Fedorenko. 2021.
\newblock The neural architecture of language: Integrative modeling converges
  on predictive processing.
\newblock \emph{Proceedings of the National Academy of Sciences},
  118(45):e2105646118.

\bibitem[{Shain et~al.(2020)Shain, Blank, van Schijndel, Schuler, and
  Fedorenko}]{shain2020fmri}
Cory Shain, Idan~Asher Blank, Marten van Schijndel, William Schuler, and
  Evelina Fedorenko. 2020.
\newblock fmri reveals language-specific predictive coding during naturalistic
  sentence comprehension.
\newblock \emph{Neuropsychologia}, 138:107307.

\bibitem[{Shliazhko et~al.(2022)Shliazhko, Fenogenova, Tikhonova, Mikhailov,
  Kozlova, and Shavrina}]{shliazhko2022mgpt}
Oleh Shliazhko, Alena Fenogenova, Maria Tikhonova, Vladislav Mikhailov,
  Anastasia Kozlova, and Tatiana Shavrina. 2022.
\newblock mgpt: Few-shot learners go multilingual.
\newblock \emph{arXiv preprint arXiv:2204.07580}.

\bibitem[{Siegelman et~al.(2022)Siegelman, Schroeder, Acart{\"u}rk, Ahn,
  Alexeeva, Amenta, Bertram, Bonandrini, Brysbaert, Chernova
  et~al.}]{siegelman2022expanding}
Noam Siegelman, Sascha Schroeder, Cengiz Acart{\"u}rk, Hee-Don Ahn, Svetlana
  Alexeeva, Simona Amenta, Raymond Bertram, Rolando Bonandrini, Marc Brysbaert,
  Daria Chernova, et~al. 2022.
\newblock Expanding horizons of cross-linguistic research on reading: The
  multilingual eye-movement corpus (meco).
\newblock \emph{Behavior Research Methods}, pages 1--21.

\bibitem[{Smith and Levy(2013)}]{smith2013effect}
Nathaniel~J Smith and Roger Levy. 2013.
\newblock The effect of word predictability on reading time is logarithmic.
\newblock \emph{Cognition}, 128(3):302--319.

\bibitem[{Sun(2023)}]{sun2023optimizing}
Kun Sun. 2023.
\newblock Optimizing predictive metrics for human language comprehension.
\newblock \emph{bioRxiv}, pages 2023--09.

\bibitem[{Sun et~al.(2023{\natexlab{a}})Sun, Wang, and
  Lu}]{sun2023interpretable}
Kun Sun, Qiuying Wang, and Xiaofei Lu. 2023{\natexlab{a}}.
\newblock An interpretable measure of semantic similarity for predicting eye
  movements in reading.
\newblock \emph{Psychonomic Bulletin \& Review}, pages 1--16.

\bibitem[{Sun and Wang(2022)}]{sun2022semantic}
Kun Sun and Rong Wang. 2022.
\newblock Semantic similarity and mutual information predicting sentence
  comprehension: the case of dangling topic construction in chinese.
\newblock \emph{Journal of Cognitive Psychology}, pages 1--24.

\bibitem[{Sun et~al.(2023{\natexlab{b}})Sun, Wang, and
  Baayen}]{sun2023attention}
Kun Sun, Rong Wang, and Harald Baayen. 2023{\natexlab{b}}.
\newblock Attention-aware measures of semantic relevance for predicting human
  reading behavior.
\newblock \emph{Linguistics}.

\bibitem[{Traxler(2011)}]{traxler2011introduction}
Matthew~J Traxler. 2011.
\newblock Introduction to psycholinguistics: Understanding language science.

\bibitem[{Van~Schijndel and Linzen(2021)}]{van2021single}
Marten Van~Schijndel and Tal Linzen. 2021.
\newblock Single-stage prediction models do not explain the magnitude of
  syntactic disambiguation difficulty.
\newblock \emph{Cognitive Science}, 45(6):e12988.

\bibitem[{Vasishth and Lewis(2006)}]{vasishth2006argument}
Shravan Vasishth and Richard~L Lewis. 2006.
\newblock Argument-head distance and processing complexity: Explaining both
  locality and antilocality effects.
\newblock \emph{Language}, pages 767--794.

\bibitem[{Vaswani et~al.(2017)Vaswani, Shazeer, Parmar, Uszkoreit, Jones,
  Gomez, Kaiser, and Polosukhin}]{vaswani2017attention}
Ashish Vaswani, Noam Shazeer, Niki Parmar, Jakob Uszkoreit, Llion Jones,
  Aidan~N Gomez, {\L}ukasz Kaiser, and Illia Polosukhin. 2017.
\newblock Attention is all you need.
\newblock \emph{Advances in neural information processing systems}, 30.

\bibitem[{Venhuizen et~al.(2019)Venhuizen, Crocker, and
  Brouwer}]{venhuizen2019expectation}
Noortje~J Venhuizen, Matthew~W Crocker, and Harm Brouwer. 2019.
\newblock Expectation-based comprehension: Modeling the interaction of world
  knowledge and linguistic experience.
\newblock \emph{Discourse Processes}, 56(3):229--255.

\bibitem[{Vrieze(2012)}]{vrieze2012model}
Scott~I Vrieze. 2012.
\newblock Model selection and psychological theory: a discussion of the
  differences between the akaike information criterion (aic) and the bayesian
  information criterion (bic).
\newblock \emph{Psychological Methods}, 17(2):228.

\bibitem[{Wieling(2018)}]{wieling2018analyzing}
Martijn Wieling. 2018.
\newblock Analyzing dynamic phonetic data using generalized additive mixed
  modeling: A tutorial focusing on articulatory differences between l1 and l2
  speakers of english.
\newblock \emph{Journal of Phonetics}, 70:86--116.

\bibitem[{Wilcox et~al.(2020)Wilcox, Gauthier, Hu, Qian, and
  Levy}]{wilcox2020predictive}
Ethan~Gotlieb Wilcox, Jon Gauthier, Jennifer Hu, Peng Qian, and Roger Levy.
  2020.
\newblock On the predictive power of neural language models for human real-time
  comprehension behavior.
\newblock \emph{arXiv preprint arXiv:2006.01912}.

\bibitem[{Wolf et~al.(2020)Wolf, Debut, Sanh, Chaumond, Delangue, Moi, Cistac,
  Rault, Louf, Funtowicz et~al.}]{wolf2020transformers}
Thomas Wolf, Lysandre Debut, Victor Sanh, Julien Chaumond, Clement Delangue,
  Anthony Moi, Pierric Cistac, Tim Rault, R{\'e}mi Louf, Morgan Funtowicz,
  et~al. 2020.
\newblock Transformers: State-of-the-art natural language processing.
\newblock In \emph{Proceedings of the Conference on EMNLP: System
  Demonstrations}, pages 38--45.

\bibitem[{Wood(2017)}]{wood2017generalized}
Simon~N Wood. 2017.
\newblock \emph{Generalized Additive Models: An Introduction with R}.
\newblock Chapman and Hall/CRC.

\bibitem[{Wood(2020)}]{wood2020inference}
Simon~N Wood. 2020.
\newblock Inference and computation with generalized additive models and their
  extensions.
\newblock \emph{Test}, 29(2):307--339.

\bibitem[{Wood et~al.(2016)Wood, Pya, and S{\"a}fken}]{wood2016smoothing}
Simon~N Wood, Natalya Pya, and Benjamin S{\"a}fken. 2016.
\newblock Smoothing parameter and model selection for general smooth models.
\newblock \emph{Journal of the American Statistical Association},
  111(516):1548--1563.

\end{thebibliography}
\bibliographystyle{aclnatbib}

\appendix
\onecolumn
\section*{Appendices}

\subsection*{A. Differences between reading speed and fixation duration}

In reading research, reading speed and fixation duration for individual words represent distinct but related aspects of how we process language. The following specifies these differences.

Reading speed (or average reading speed, or reading rate) is typically measured in words per minute or second (e.g., wpm) and reflects the overall rate at which a person reads a passage of text or a sentence. Reading speed is usually employing text or individual sentence as the unit of interest. This metric gives a general indication of reading efficiency or fluency. It measures the reader's ability to comprehend and process text or sentence over a given period. This can vary significantly depending on the reader's skill, the text's complexity, and the reading purpose (e.g., reading for pleasure vs. critical reading). Reading speed encompasses the entire reading process, including both the time spent on fixations (when the eyes stop on a word) and saccades (the rapid movements between fixations), as well as regressions (backward movements to re-read text) and pauses for comprehension. Reading speed has been extensively and intensively explored (\citealp{jackson1979processing}; \citealp{herman1985effect}; \citealp{carver1990reading}; \citealp{rayner2010eye}).  

Fixation duration for individual words (here total fixation duration) refers to the total length of time the eyes remain stationary during fixations on a word while reading. These durations are measured in milliseconds with an eye-tracker and can indicate the cognitive effort required to process the word. Fixation duration on individual words can reveal how word characteristics (such as length, frequency, and predictability) and contextual difficulty affect reading. Longer fixations tend to suggest that a word is harder to recognize or comprehend within its context. Fixation durations vary across words within a text and among different readers. They provide insight into the moment-by-moment cognitive processes involved in lexical reading, such as lexical access, parsing, and integration of information.

Reading speed offers a macro-level view of reading behavior, indicating overall efficiency or difficulty at the sentence or text level. In contrast, fixation duration provides a micro-level perspective, revealing the cognitive processing time for individual words. While both metrics can be affected by the text's difficulty and the reader's proficiency, reading speed could be also influenced by broader factors like braoder context and the reader's strategy or purpose. Fixation duration is more directly related to immediate cognitive demands for word-specific characteristics.

%Understanding both metrics is crucial for a comprehensive picture of reading behavior. They complement each other, with average reading speed providing a broad measure of reading efficiency and fixation durations offering detailed insights into the cognitive processes underlying reading.

%Researchers use average reading speed to assess and compare reading abilities across populations and contexts. Fixation duration is often used in detailed studies of reading comprehension and processing to understand how readers interact with specific linguistic or textual features.

\subsection*{B. Computing sentence surprisal}
\label{surprisal}
This section details how to compute sentence-level surprisal using the three methods: chain rule (CR), negative log-likelihood (NLL) and next sentence prediction (NSP). The first two methods were applied in either \texttt{m-BERT} or \texttt{mGPT}, and the third method was merely implemented in \texttt{m-BERT}.

The following provides a detailed account of how to apply \textbf{chain rule (CR)} to compute sentence surprisal . First, we tokenized sentence into various tokens: $S = [t_1, t_2, \ldots, t_n]$. Probability of a token $t_i$ in the sentence \textit{S} given its left context \textit{C} in a text and preceding tokens can be represented as: $P(t_i | C, t_1, t_2, \ldots, t_{i-1})$. (\textit{S} = sentence, \textit{t} = token, \textit{C} = context). Based on this, we can calculate the probability of each token prior to its left context. After obtaining the probability of each word, we computed the joint probability of the entire sentence in equation (2):  \textit{S} given its context \textit{C}, $P(S | C) = P(t_1 | C) \cdot P(t_2 | C, t_1) \cdot P(t_3 | C, t_1, t_2) \cdot \ldots \cdot P(t_n | C, t_1, t_2, \ldots, t_{n-1}) \quad (2)$, further getting sentence surprisal, $-log(P(S|C))$. The equation is an application of the chain rule of probability to sequences (like sentences). It captures the idea that the probability of a sentence given a context can be decomposed into the product of conditional probabilities of its individual tokens. The following provides the details why it works. When you have a sequence of events or tokens, the chain rule can be extended. For a sequence of three events $t_1$, $t_2$, and $t_3$: $P(t_1, t_2, t_3) = P(t_1) \times P(t_2 | t_1) \times P(t_3 | t_1, t_2)$. In the case of natural language, sentences often have a context. The context can be prior knowledge, a preceding sentence, or any other relevant information. Given a context (\textit{C}), the probability of a sequence changes (\textit{S}): $P(S | C) = P(t_1 | C) \times P(t_2 | C, t_1) \times P(t_3 | C, t_1, t_2) \times \ldots \times  P(t_n | C, t_1 , t_2 , ..., t_{n-1} )$ (i.e., the \textbf{multiplication} of probabilities of each token given the left context within a sentence). The method of computing sentence probability is consistent with the principles of probability theory and the way sequence modeling is approached in the context of NLP. 
%It allows us to model the dependencies between tokens in a sequence, which is essential in natural language where the likelihood of a word can depend heavily on the words that precede it. %The chain rule of probability can break down the joint probability of a sequence into a product of conditional probabilities, that is, the joint probability of sentence can be calculated as the product of the conditional probabilities of each event (token) given the previous events (tokens).
%We normalized the sentence probability by dividing the total words in a sentence: $\text{Normalized} P(S | C) = \frac{P(S | C)}{n}$, and the normalized sentence surprisal can be obtained, $-log(\text{Normalized} P(S | C))$. Finally, we used the normalized sentence surprisal as a metric. 

Moreover, using \textbf{Negative Log-Likelihood (NLL)}, we can also compute the surprisal of a sentence conditioned on its preceding textual content. The specific procedure is detailed as follows: the text is tokenized into a sequence of token IDs, converted into a PyTorch tensor, and processed through the model to determine the NLL. NLL is used as a loss function to measure
how well the model's predictions align with the actual data. For a given sequence of tokens, NLL is a measure of how surprised the model is by the actual sequence. There are two steps in calculation: 1) \textbf{Sum up} these NLLs for the words in the sentence: $\text{NLL} (S \big| C) = −\sum \log(P(t_i | t_1 , t_2 , ..., t_{i-1})$ ($t_i$ is the word in the target sentence ``S'', and $t_1$ , $t_2$ , ..., $t_{i-1}$ include both the context and the preceding words in the sentence)
2) Convert NLL to Probability, shown in equation (3): $P(S | C) = e^{-\text{NLL}(S | C)} \quad (3)$. The term $e^{-\text{NLL}}$ represents an exponential function with the negative Natural Log Loss (NLL) of $S$ given $C$ as the exponent.

The following elaborates on the \textbf{differences} between applying Chain Rule (\textbf{CR}) and Negative Log Likelihood (\textbf{NLL}) within the contexts of \textbf{BERT} and \textbf{GPT}. We detail these distinctions specifically in the scenarios of BERT and GPT, respectively.

Within the context of autoregressive language models (LMs) like \texttt{GPT}, the concepts of the CR for probability and NLL indeed converge when computing sentence surprisal. The CR decomposes the probability of a sequence into the product of conditional probabilities. This method aligns with how autoregressive LMs generate text, predicting each token based on the previous ones. NLL is often used as a loss function in training these models, calculated by taking the negative logarithm of the probability of the observed data under the model. For sentence, NLL would be the \textbf{sum} of the negative logs of these conditional probabilities. When you take the logarithm of the product of conditional probabilities (as per the CR), it becomes a \textbf{multiplication} of these probabilities. The math equations for CR and NLL introduced at the start of this section are applicable to the GPT case.

 However, in the context of \texttt{BERT}, we need to make slight changes because BERT is fundamentally different from autoregressive models in that it does not generate text sequentially. BERT's architecture is designed to understand the context of a token by looking at the tokens that come before and after it—this is what makes it ``bidirectional''. It is trained primarily through tasks like Masked Language Modeling (MLM), where it predicts the identity of tokens that have been artificially masked in the input text, based on the context provided by the non-masked tokens around them. The chain rule (CR), which breaks down the probability of a sequence into the product of conditional probabilities, naturally applies to models that generate or predict sequences in a linear, step-by-step manner (e.g., predicting the next word given the previous words in a sentence). While BERT does not compute sentence surprisal through the chain rule in the same way as autoregressive models, we can use BERT to estimate a form of surprisal or predictability for words in a sentence based on its bidirectional context. For instance, by masking a word in a sentence and using BERT to predict its likelihood, one can infer a measure of how predictable that word is given its context.  BERT's capacity to predict the probability of masked tokens in context allows for an indirect assessment of word-level surprisal. This approach leverages BERT's contextual predictions to gauge how expected or surprising a word is within its sentence context. Assuming a sequence of tokens S = [$t_1$, $t_2$, \ldots, $t_n$], and a context C, we conceptually mask and predict each token $t_i$ based on all other tokens as context, mimicking a sequential dependence that does not naturally exist in BERT. The ``conditional probability'' of each token $t_i$, given its context (both preceding and succeeding, \textit{C}), is estimated as $ P(t_i |C, t_{\text{rest}})$, where $t_rest$ represents the rest of the tokens serving as context which consists the target sentence and the previous stentences. Sentence surprisal would then be represented as Equation (4): \[
 \text{Surprisal}(S|C) = -\log \prod_{i=1}^{n} P(t_i | C, t_{\text{rest}}) \qquad (5)
 \]. 
 
 We then focus on NLL in \texttt{BERT}. While NLL is a concept that can be applied broadly in machine learning to measure how well a model's predictions match the actual data (and is used as a loss function in many contexts, including training BERT), its specific application and interpretation can vary. In the context of BERT, NLL is used to optimize the model's ability to accurately predict the masked tokens based on their context, which involves both preceding and succeeding tokens in a sentence. In other words, BERT predicts the identity of masked tokens based on their surrounding (bidirectional) context, and NLL quantifies the model's prediction accuracy. %The use of NLL in BERT is to minimize the NLL, thereby improving the model's predictive performance. 
 For each masked token prediction, compute the NLL as:
 $\text{NLL}(S|C) = -\sum \log(P(t_i|t_1, t_2, ..., t_{i-1}, t_{i+1}, ..., t_n))$
 Here, $t_i$ is the masked token, and the context $C$ includes both preceding and succeeding tokens. \textbf{Sum up} the NLLs for all masked tokens in the sentence to get the total NLL, representing the sentence surprisal. %This is a different operational context compared to how NLL is conceptualized in sequential, token-by-token prediction models (e.g., GPT).   % In BERT, computing sentence surprisal isn't straightforward because the model doesn't inherently generate text in a linear fashion where one token's probability is conditioned strictly on the preceding tokens. Instead, BERT evaluates tokens in a broader context, which changes how surprisal would be calculated or interpreted.

 %This is precisely what the NLL calculation does. %The only difference is the perspective: CR calculates the probability of a sequence directly, while NLL frames it in terms of surprisal or uncertainty from the model's viewpoint. %In the end, both approaches measure how predictable a sentence is given its context, with lower probabilities (or higher NLLs) indicating more surprisal. In the context of sentence surprisal calculation, using CR or NLL with GPT should theoretically yield the same result because both are based on the same underlying probabilities. For autoregressive LMs, the process of next-token prediction inherently follows the chain rule by considering each token's probability conditioned on its predecessors. The negative log-likelihood, used to measure model performance and train these models, is mathematically related to the chain rule through the transformation of multiplication of probabilities into a sum of their logarithms. 
%In this sense, while CR and NLL can be described differently—in terms of probability theory versus model training and evaluation—their application to computing sentence surprisal in the context of autoregressive language models converges to the same fundamental principles and outcomes.

%How Chain Rule and NLL Converge in Autoregressive LMs:
%Chain Rule (CR): The CR decomposes the probability of a sequence into the product of conditional probabilities

In addition to adopting the chain rule and NLL, we could employ the \textbf{``next sentence prediction (NSP)''} mechanism in BERT to compute sentence surprisal. To compute the probability that one sentence follows another using BERT's NSP, begin by preparing the input. Combine the two sentences, placing a special [CLS] token at the start and a [SEP] token in between them. After tokenizing this combined sequence with BERT's tokenizer, pass it through the BERT model. BERT uses the representation of the [CLS] token, which encapsulates information about the entire input, to predict the relationship between the two sentences. The model outputs two probabilities: ``IsNext'' and ``NotNext''. The ``IsNext'' probability indicates the likelihood that the second sentence logically follows the first. By examining this probability, one can gauge how likely the model perceives the given sentence order to be. The above description can be summarized as the following two equations. Probability of sentence B being the next sentence given the left context A is formalized as (6):

$$P(B|A) = \text{softmax}(w_2^T [\text{CLS}] + b_2)\qquad (6)$$

where [\text{CLS}] denotes the representation of the first sentence obtained from the final layer of the BERT model, and the softmax function computes the probability distribution over all possible next sentences. $w_2$ and $b_2$ are learnable weight vector and bias term, respectively. And then we can obtain the sentence surprisal. However, NSP in BERT is a binary classifier, that is, the model determines if a sentence logically follows a given sentence or not. This is a simplification of the rich structure and semantics in natural language. The use of NSP may raise the question about the interpretability of BERT's estimated probabilities for subsequent sentences, especially when viewed from a cognitive modeling standpoint. The reason for this is that we probably argue that as humans read, they probably take more than two sentences as the context window. Despite this, we still need to apply the statistical analysis to test whether such metrics are useful or not.   

\subsection*{C. BERT approximates human reading behavior}
Some researchers have underestimated BERT's potential in calculating word probabilities, arguing that the use of BERT (masked LMs) is hard to justify within the realm of cognitive modeling. Their concern stems from the fact that a masked language model like \texttt{m-BERT} can consider context from both directions, including rightward context, which could not be typically how human comprehension works. However, BERT is better able to estimate word probability because it can be really line with human reading behavior. The following details this point.   

Those familiar with eye-tracking in reading recognize the preview benefit (i.e., parafoveal-on-foveal effects, ``PoF'' for short). Parafoveal-on-foveal effects in reading refer to the influence of information from the words located in the parafoveal region (the area immediately surrounding the fixated word) on the processing of the currently fixated word (the foveal word) (\citealp{kennedy2005parafoveal}; \citealp{kliegl2007preview}; \citealp{schotter2012parafoveal}; \citealp{hohenstein2014semantic}). As the parafoveal-on-foveal effect takes place in the context of reading and visual processing, the visual perception of words not directly fixated upon (parafoveal words) influences the processing of the word currently fixated on (the foveal word).

\begin{wrapfigure}{r}{0.5\textwidth} 	
	\centering
	
	\includegraphics[width=0.45\textwidth]{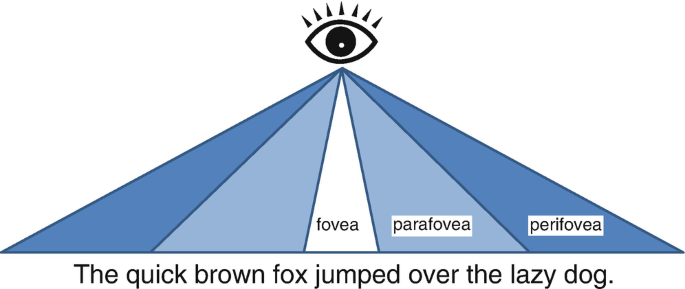}
	
	\caption{The parafoveal-on-foveal effects in reading (from \citet{Masato2023polychromy})}
	
	\label{fig:parafovea}
\end{wrapfigure}

Put it simply,  when we read, our eyes do not move smoothly across the text but make quick, jerky movements known as saccades, interspersed with brief pauses called fixations. During these fixations, the eye directly focuses on a small area of text. This area, where visual acuity is highest, is known as the foveal region. Surrounding the foveal region is the parafoveal region, where vision is not as sharp but still capable of processing some information about the text, such as word length or initial letters. The PoF effect refers to how information from the parafoveal region can pre-activate or facilitate the processing of the word once it becomes the focus of the next fixation (moves into the foveal region), thereby affecting reading speed and comprehension. For example, if the parafoveal word is semantically or syntactically related to the foveal word, it might speed up the recognition and processing of the foveal word when the eye moves to it. In short, processing the target word actually is involved in incorporating the information on the next one or two ore more words (n+1, or n+2 word in literature) through previewing, as shown in Fig. \ref{fig:parafovea}. 

%Understanding the PoF effect is crucial for developing theories of eye movement control and visual processing in reading, as it sheds light on the complex interplay between visual attention, lexical processing, and the mechanics of eye movements.

Relevant research shows that reading times are shorter for a target word when it matches the preview word, compared to when they are different. This suggests that the preview word is processed using parafoveal vision. BERT mirrors this process to some extent because it can incorporate information from subsequent words, approximating how human parafoveal vision previews words to the right of the fixation. In this sense, BERT aligns with human reading patterns to a greater degree. The surprisal values calculated by BERT, reflecting real-world reading dynamics, potentially provide more precise predictions of word reading times. The criticism that BERT generates only pseudo-surprisal overlooks its relevance to real-world human reading behaviors. On the other hand, surprisal values computed by GPT do not account for the previewing aspect of reading. Despite this, surprisal values from both GPT and BERT can, in theory, offer insights into reading behaviors. Nevertheless, statistical analysis is needed to determine which model's surprisal predictions align more closely with observed reading data.

\subsection*{D. Attention-aware approach and its memory capability}
\label{attention}
This section provides a comprehensive guide on calculating sentence relevance using an attention-aware approach, executed in a two-step strategy. The sliding window includes four sentences (t, 2, 1, and n1), as shown in the {\color{blue}{Panel B}} of Fig. \ref{fig:computation}.  ``t'' represents the target sentence, and ``n1'' for the next sentence. We aim to compute how the target sentence (t) is semantically related with the other three sentences (2, 1, and n1 form the context window).   %By adding the semantic relevance \footnote{Either \textit{cosine} similarities or \textit{correlations} among word embeddings serve as a computational method to calculate the semantic relevance between two words.

The initial step involves generating sentence embeddings using \texttt{m-BERT} or \texttt{mGPT}, followed by computing the similarity between two sentences based on these embeddings. The subsequent step entails applying the ``attention-aware'' approach to manage multiple similarity values across several sentences within a window stack.

In the first step, we need to generate embeddings for each sentence in this window by employing BERT or GPT. After obtaining sentence embedding for each sentence, we applied \texttt{cosine similarity} to compare the sentence embeddings generated by \texttt{m-BERT} or \texttt{mPGT}, and this practice is a common approach for computing semantic similarity between sentences, formalized as Equation (7).
$$\text{similarity}(s,c) = \frac{e_s \cdot e_c}{\lVert e_s \rVert \lVert e_c \rVert} \qquad (7)$$
where $s$ is the input sentence, $c$ is the left context, $e_s$ and $e_c$ are their respective sentence embeddings obtained using mean pooling with BERT, $\cdot$ denotes dot product, and $\lVert \cdot \rVert$ denotes L2 norm. The second term is a modified form of the cosine similarity to account for the distance between the embeddings.

We employed two distinct multilingual LLMs (i.e., \texttt{m-BERT} and \texttt{mGPT}), noting subtle differences in their approaches to generating sentence embeddings. Regarding \texttt{m-BERT}, when a sentence is input into BERT, it is first tokenized and then prepended with a special ``\texttt{[CLS]}'' token. After processing through BERT's layers, the embedding corresponding to this ``\texttt{[CLS]}'' token is often used as the sentence embedding. The tokenized input sentence (S) is represented as \texttt{[CLS]}, $t_1$, $t_2$, \ldots, $t_n$. After processing through BERT, the output embeddings at the final layer for this sequence are [$E_{\text{CLS}}$, $E_{t_1}$, $E_{t_2}$, \ldots, $E_{t_n}$]. Mean pooling involves calculating the average of the embeddings of all tokens in the sequence \texttt{[CLS]}.

Here is Equation (8) to compute a sentence (or a text) embedding using \texttt{m-BERT}.
%\begin{equation}
$$	e_s = \text{BERT}_{\text{pooler}}(s) = \frac{1}{n}\sum_{i=1}^n h_i \quad (8)$$
%\end{equation}
where $s$ is the input sentence, and $\text{BERT}_{\text{pooler}}$ is a mean pooling layer that takes the output vectors $h_i$ of all $n$ tokens in the input sentence and computes their average to obtain the sentence representation $e_s$.

Moreover, we used \texttt{mGPT} to compute sentence similarity for cross-validation. We still applied embedding-based method to do this. Specifically, \texttt{mGPT} was to obtain embeddings for each sentence. Still employing the mean pooling, we used the hidden states of the sentence to represent the embedding. Cosine similarity was to calculate the similarity between the two embeddings, which is similar to BERT-similarity computation. 

The subsequent discussion highlights the distinctions between \texttt{m-BERT} and \texttt{mGPT} in generating sentence-based embeddings. First, \texttt{mGPT}, being primarily focused on generative tasks, does not utilize a special token like \texttt{[CLS]} for aggregating sentence meaning. GPT architecture is designed to predict the next token in a sequence based on the previous context, which inherently focuses on a unidirectional flow of information. The other difference is to use pooling for sentence embeddings. Specifically, to obtain sentence-level embeddings from GPT, one common method is to aggregate the hidden states (from the last layer) of all tokens in the output. Mean or max pooling can be applied to these token-level embeddings to create a single vector representing the entire sentence. This approach leverages the contextual information encoded by GPT in a sequential manner, albeit without the bidirectional context that BERT captures. In BERT, pooling is an alternative to using the ``\texttt{[CLS]}'' token embedding, offering a way to capture a distributed representation of sentence meaning. However, in GPT, pooling is a necessary step for sentence-level representation since the model lacks a mechanism like the \texttt{[CLS]} token for summarizing the text.

%After finishing the first step, we detail how to use the attention-aware approach to calcuate sentence revelcance. After obtaining the similarity values for the sentences in the window stack, we applied the weights to process these similarity values and add the weighted values to get a final value, which has been discussed in the main body. We want to emphase why the attention-aware approach can work to capture the contextual information and the underlying mechanin of the weigths. 
The similarity score between any two sentences in this window is calculated. Following this, four cosine similarity values are obtained for the sentences. Subsequently, we implemented the second step.

After completing the first step, we elaborated on applying weights to calculate sentence relevance. Upon acquiring the similarity values for sentences within the window stack, we proceeded to apply weights to these values. The process involves aggregating the weighted similarity values to derive a final score, as detailed in the main text. Our aim is to underscore the efficacy of the attention-aware approach in capturing contextual information and to explain the underlying mechanism of the weighting system.

As illustrated in the {\color{blue}{Panel B}} of Fig. \ref{fig:computation}, in a window stack, sentences closer to the target sentence resemble the initial days in the forgetting curve, while more distant words resemble the latter days. To simulate human forgetting mechanism, we allocated larger weights to the closer sentences and smaller weights to the distant sentences and the similarity between human forgetting mechanism and attentional weights adopted in the current study. The weight values gradually decrease with the distance between the target sentence and the contextual sentences (see Equation (1) and the Panel B of Fig. \ref{fig:computation}), similar to the forgetting curve \citep{loftus1985evaluating}.  The attention-aware approach could be linked to memory models in terms of how memory is decayed during the encoding of information, which subsequently affects how humans process sentences during reading. More importantly, the attention-aware approach is computational and fundamentally memory-based, facilitating memory storage, retrieval, and integration. This approach not only realizes the memory function but also incorporates the expectation effect.

Furthermore, when applying a 3-sentence window (the target sentence is not takne into account), akin to the fading of memories over three days, the average rate of memory decline can be estimated to average one-third per day. This fading initiates more abruptly and slows as time progresses. By modeling this, we can recalibrate the significance of sentences based on their distance from the target sentence, diminishing their value by a one-third. In this sense, sentences further from the target sentence receive lower weights compared to those in a uniform distribution. This leads to a gradual weighting scale, such as 1/3, 2/3, 1/2, and so forth, mirroring their relative proximity to the target sentence.

\begin{figure*}[!h]%[htp]
	
	\centering
	
	\includegraphics[width=0.82\textwidth]{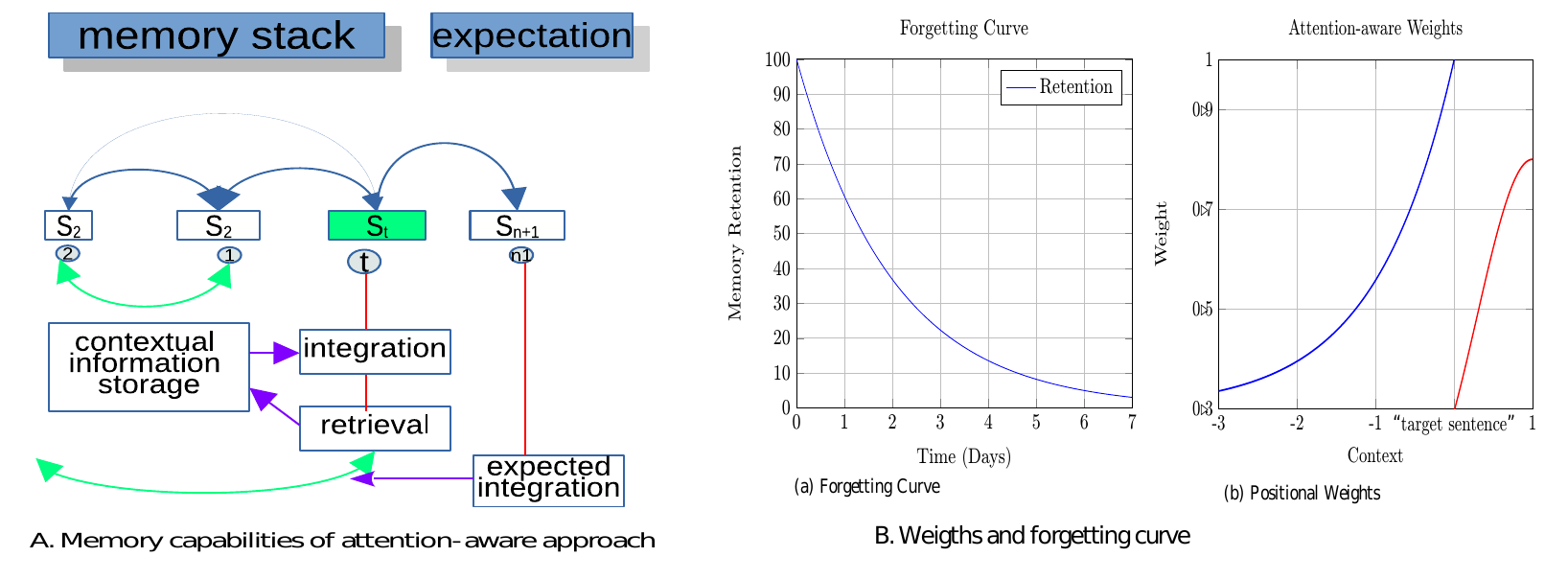}
	
	\caption{The memory capability and weights adopted in the attention-aware approach}
	
	\label{fig:memory}
	
\end{figure*}

The {\color{blue}{Panel A}} of Fig.\ref{fig:memory} shows how the attention-aware approach we adopted simulates a short-term memory stack, mirroring how readers retain memory of previously encountered words and their meanings. The method for using various weights of semantic relevance between any two sentences is inspired by both the attention mechanism found in Transformers ({\color{blue}{Panel A}} in Fig.\ref{fig:memory}) and the human process of forgetting ({\color{blue}{Panel B}} in Fig. \ref{fig:memory}). The attention-aware approach can facilitate effective incorporation of contextual information and enabling it to achieve memory storage, retrieval, and integration. %The weight values gradually decrease with the distance between the target sentence and the contextual sentences (see Equation (1) and the Panel B of Fig. \ref{fig:computation}), similar to the forgetting curve \citep{loftus1985evaluating}. 

Finally, as previously noted, the term ``attention-aware'' bears a resemblance to the algorithm of attention in Transformers, primarily because the attention-aware approach we proposed significantly enhances computational efficiency through the incorporation of contextual information. However, our attention-aware method does not incorporate any attention layers from Transformers. Each step in the computation of attention-aware metrics is transparent and interpretable.

\subsection*{E. Statistical methods and comparison standards}
\label{statistcal}
To meet our goals of accurately predicting multilingual eye-tracking data, we utilized Generalized Additive Mixed Models (GAMMs)\citep{wood2017generalized}. GAMMs are effective in analyzing nonlinear effects and multiplicative interactions between variables, making them ideal for evaluating the predictability of semantic similarity. They are more flexible than traditional regression methods in modeling complex relationships between variables. Eye-tracking data is simpler to analyze statistically than EEG and fMRI data, which makes it an ideal choice for our study on naturalistic discourse reading. However, assessing model performance and comparing models can be challenging, and relying solely on correlations can be limiting. Fortunately, GAMMs are well-suited for comprehensive and precise assessments of model performance. We compared models using \texttt{AIC} (Akaike's Information Criterion) values, where a smaller value indicates a better model.

\texttt{AIC} or \texttt{BIC} (Bayesian Information Criterion ) are both measures of model fit that balance goodness of fit with model complexity. Lower values of \texttt{AIC} or \texttt{BIC} indicate better model fit. However, \texttt{AIC} is a popular criterion for comparing GAMMs, and it has some advantages over other criteria. \texttt{AIC} is designed to balance the trade-off between model fit and model complexity, penalizing models with more parameters. This makes it useful for selecting models that provide a good balance between fit and complexity. \texttt{AIC} is also relatively easy to compute and widely used in statistical modeling.
 
 Comparing two GAMM (or LMER, Linear Mixed Effects Models) models, where one is larger than the other (having all the parameters of the other model and some additional ones), the likelihood will always be higher for the larger model. This is because a larger model can fit the data better by having more parameters. However, directly comparing loglikelihoods is not appropriate when models have different sizes. Both \texttt{AIC} and \texttt{BIC} address this issue by incorporating a penalty for the number of parameters. \texttt{AIC} and \texttt{BIC} have different principles for penalizing complexity. Generally, \texttt{BIC} penalizes complexity more strongly than \texttt{AIC}, tending to favor smaller models unless both approaches agree. In essence, \texttt{AIC} is preferable when the main goal is prediction quality, as a slightly larger model can still provide good predictions, while a too small model usually does not. On the other hand, \texttt{BIC} aims to identify a reasonably sized true model by prioritizing parsimony. \texttt{BIC} is often better in finding the true model, but it has a higher chance of selecting a model that is too small, which is not favorable for prediction. In practice, the true model is often not ``small'',  but for reasons such as interpretability, smaller models are sometimes preferred even if they have slightly worse prediction performance, in which case \texttt{BIC} may be preferred. In summary, from the perspective of \texttt{AIC}, it is better to fit a slightly larger model than a too small one for improved prediction quality. However, from the perspective of \texttt{BIC}, both excessively large and excessively small models are equally undesirable \citep{vrieze2012model}.
 
 Moreover, the developer of R package on GAMM (``mgcv'') used AIC to make model comparison (\citealp{wood2016smoothing}; \citealp{wood2020inference}). AIC has also been mostly taken to understand model performance in psycholinguistic research (\citealp{wieling2018analyzing}; \citealp{baayen2020introduction} and the relevant studies) if GAMM or generalized mixed-effect models are employed.
 
 In the studies conducted by \citet{wilcox2020predictive} and \citet{oh2023does}, the relationship between model perplexity and \texttt{$\Delta$LogLik} (log-likelihood) was utilized to analyze the perceptual competence of surprisal generated by different LMs. Their objective was to determine which LMs were capable of generating more powerful surprisal based on various corpora. In contrast, the current study aims to assess the predictive performance of our algorithms. 
 
 We should point out some potential issues when using \texttt{$\Delta$LogLik} as a criterion. \citet{wilcox2020predictive} utilized GAMMs for their analysis but did not incorporate any \textbf{random variables} in these models. Consequently, confirming the optimality of these GAMMs becomes challenging. On the other hand, \citet{oh2023does} mentioned that they employed LMER and included random effects. However, the LMERs used in \citet{oh2023does} did not include ``word frequency'' as a control predictor. It is well-established in psycholinguistics and cognitive science that both ``word length'' and ``word frequency'' are significant variables in predicting reading time. Including these two factors as \textbf{control predictors} is commonly practiced when studying reading time. Therefore, when investigating reading time, it is advisable to include both ``word length'' and ``word frequency'' in GAMMs or LMERs. Meanwhile, an optimal model should also include random variables. Failure to do so may result in suboptimal GAMMs or LMERs for studying reading time. Moreover, The excessive application of heavy penalties (e.g., \texttt{k} in GAMMs) on the given metrics leads to overfitting in the mixed-effect models. For example, the heavy penalty on the specific metric results in the partial effect curve of this given metric becoming much steeper, and the \texttt{AIC} or likelihood in the model increasing remarkably. Conversely, removing such penalties eliminates these effects.  %However, the LMERs used in   \citet{oh2023does} did not include ``word frequency'' as a control predictor. ``Word length'' and ``word frequency'' have been extensively investigated to show that the two variables are significant in predicting reading time, which is common sense in psycholinguistics and cognitive science. In most cases, the two factors are taken as control predictors when studying reading time, in other words, the two factors should be included in GAMMs or LMERs. Otherwise, the GAMMs or LMERs for investigating reading time are unlikely to be optimal.  Having clearly defined optimal LMERs or GAMMs is crucial to ensure accurate evaluation of computational models. Without optimization, there is a risk of encountering misleading issues that can undermine the validity of the results. %It seems that in the field of computational linguistics and NLP, there is often a lack of emphasis on optimizing regression models when employing LMERs or GAMMs. This stands in contrast to psycholinguistic research, where these statistical models are carefully utilized to gain a thorough understanding of the relationships among variables. It requires expertise and patience to achieve optimal LMER or GAMM models. 
%Moreover, $\Delta$LogLik may not be a suitable criterion for regression model comparison, which has been detailed above.
 %\citet{wilcox2020predictive} and \citet{oh2023does} used the relation between model complexity and $\triangle$LogLik to help analyze the perpection comptenece of surprisal generated by different LMs. Their purpose is to know which LMs could generate more powerful LMs based on different corpora. By contrast, the current study is to know our algorithms could work on prediction based on the same corpus. That is why the current study employed GAMM. We have to point out some potential problems in using $\triangle$LogLik.  \citet{wilcox2020predictive} employed GAMM models to produce \citet{wilcox2020predictive}, but these GAMM models did not include any random variables. In this case, it is very difficult to confirm that these GAMMs are optimal. \citet{oh2023does} mentioned that they employed LMER and also included random effects. However, $\triangle$LogLik may not be a good criterion to make model comparison.

Considering these factors, the GAMM models used in the present study include two control predictors: the mean word length for a sentence and the mean word length for a sentence. Additionally,  random variables, participants in eye-tracking experiments, and languages, were included. We used \texttt{fs} (i.e.,random smooths) to adjust the trend of a numeric predictor in a nonlinear way, which includes random intercept and random slope. In other words, we can allow the metrics of interest to explore their effects on various levels of random variables comprehensively. We employed \texttt{AIC} to compare the performance of different GAMMs. The baseline model excludes the main predictor of interest (sentence surprisal or attention-aware sentence surprisal) but retains the other elements in the full GAMM model. After fitting the regression models, we calculated the \texttt{$\Delta$}AIC values for each GAMM model by subtracting the \texttt{AIC} of the base GAMM model from that of a full GAMM model. A smaller \texttt{AIC} indicates better model performance. Similarly, a smaller \texttt{$\Delta$}AIC also indicates better performance. Additioanlly, considering the best sentence surprisal and sentence relevance in each language, we used \texttt{t} tests to check wether \texttt{$\Delta$}AIC values for two metrics have significant differences. The result shows that there is no significance difference. In other words, it is not easy to distinguish sentence surprisal or sentence relevance could predict reading speed better or not.

\subsection*{F. Performance of individual languages}

\begin{figure*}[htp]%[htp]
	
	\centering
	
	\includegraphics[width=1\textwidth]{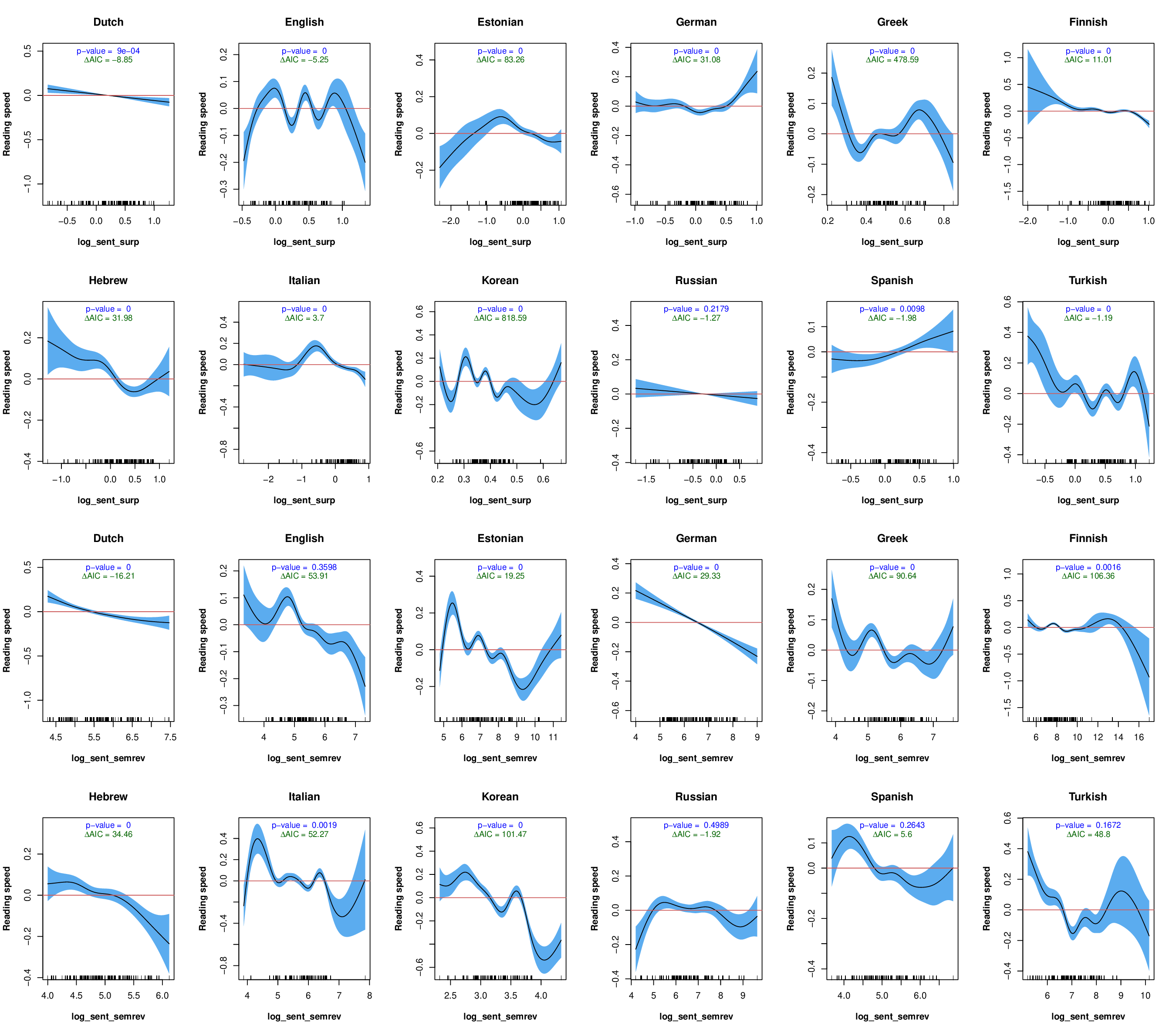}
	
	\caption{The partial effects of the primary predictors—sentence surprisal and sentence relevance—on reading speed across 13 languages (i.e., Dutch, Estonian, Finnish, German, Greek, Hebrew, Italian, Korean, Norwegian, Russian, Spanish, Turkish). Note: The upper section of the diagram features sentence surprisal, while the lower portion is dedicated to sentence relevance.The \textit{x-axis} signifies the computational metric, while the \textit{y-axis} delineates the reading speed. To achieve a closer approximation to a normal distribution, and consequently improve the fitting, all metrics undergo a logarithmic transformation. Each curve visually articulates the correlation between a predictor variable and the response variable, namely reading speed. A steeper incline on these curves underscores a more robust impact between the predictor and reading speed, whereas gentler slopes imply a less pronounced effect. Moreover, when a curve fluctuates around zero, its effect vanishes. The information regarding \textit{p}-values and $\Delta$AIC is displayed at the top of each plot. The methodology for calculating $\Delta$AIC for ``sentence surprisal'' and ``sentence relevance'' is detailed in the main text. In conclusion, sentence surprisal seems to lack significance in English, Korean, and Russian. However, sentence relevance may show no significant impact in English, Russian, Spanish, and Turkish.}
	
	\label{fig:effects}
	
\end{figure*}

The performance of an individual language is shown in Fig. \ref{fig:effects}. There are two key standards for interpreting the curve of partial effect in each plot. The first standard involves analyzing the curve's steepness. A steeper incline signifies a stronger correlation between the predictor and reading speed, whereas a gentler slope indicates a less pronounced effect. The second standard focuses on the curve's fluctuation around zero; a curve that hovers around zero suggests its impact is minimal. For instance, in the English plot, a curve consistently near zero would indicate a weak effect. Understanding mixed-effect models is essential for appreciating the significance of these indicators, highlighting the Akaike Information Criterion (AIC) as the preferred tool for model comparison.

As shown in Fig. \ref{fig:effects}, it appears that sentence surprisal lacks significance in English, Korean, and Russian, while sentence relevance shows no significant impact in English, Russian, Spanish, and Turkish.

 Next, we employed \texttt{T-test} to assess the statistical significance of the \texttt{$\Delta$}AIC values for both sentence surprisal and sentence relevance. The analyses revealed no significant differences, with the \textit{p}-value substantially exceeding 0.05. This outcome suggests that sentence surprisal and sentence relevance may contribute equivalently to the prediction of reading speeds. As a result, it can be inferred that both metrics are effective in facilitating the overall comprehension of sentences. Despite this, Table \ref{tab:aic} shows that the overall predictive power of sentence surprisal with \texttt{m-BERT} and CR is stronger than sentence relevance.

\subsection*{G. Correlation between sentence surprisal and sentence relevance}
\label{corr}
\begin{figure}[htp]%[htp]

	\centering

	\includegraphics[width=\textwidth]{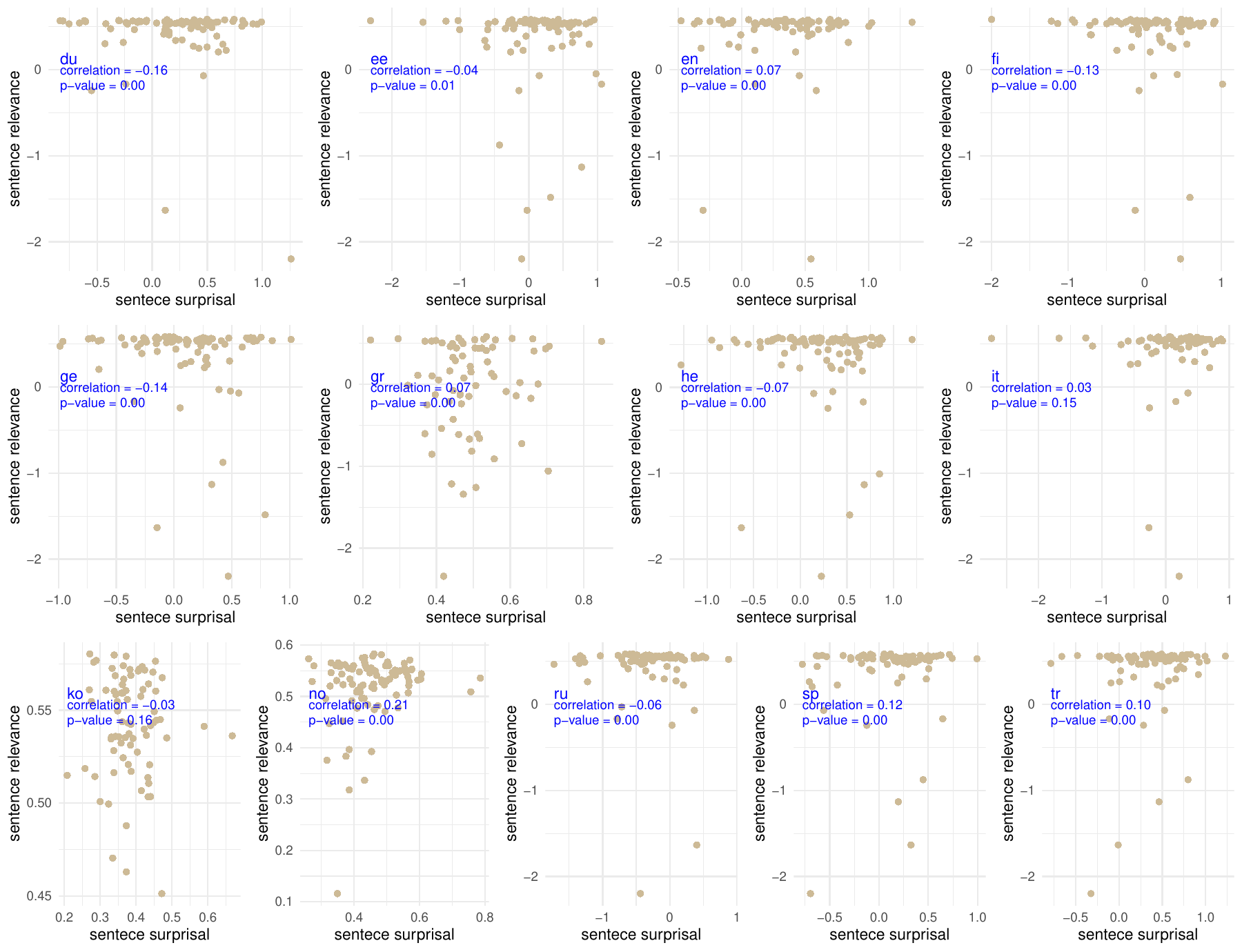}

	\caption{Pearson correlation between sentence-level surprisal (computed by \texttt{m-BERT} and chain rule) and sentence-level semantic relevance (computed based on \texttt{m-BERT}) in each language. Note: the abbreviations for these 13 melange are as follows. du = Dutch; ee = Estonian; en = English; fi = Finnish; ge = German; gr = Greek; he = Hebrew; it = Italian; ko = Korean; no = Norwegian; ru = Russian; sp = Spanish; tr = Turkish}

	\label{fig:corr1}

\end{figure}

The overall Pearson correlation between sentence surprisal (computed using \texttt{m-BERT} with the chain rule) and sentence relevance (also derived from \texttt{m-BERT}) stands at \textbf{-0.054}. The value suggests that there is a weak correlation between the two metrics, and indicating that the two metrics are completely distinct. This relationship across the 13 languages is depicted in Fig. \ref{fig:corr1}. The observed correlations among the metrics for each language are notably minimal. Such low correlation scores underscore the fact that the sentence surprisal and sentence relevance we calculated represent entirely distinct metrics.
%The overall correlation between sentence surprisal (computed by \texttt{m-BERT} with chain rule) and sentence relevance (computed by \texttt{m-BERT}) is -0.054. The correlation between sentence surprisal and sentence relevance in each language is show in Fig. \ref{fig:corr}. The correlation between two metrics in each language is quite small. These correlation values indicate that sentence surprisal and sentence relevance we computed are totally distinct metrics. 

\bigskip

\end{document}